\documentclass{article}

\usepackage[preprint]{neurips_2026}

\usepackage[utf8]{inputenc} 
\usepackage[T1]{fontenc}    
\usepackage{hyperref}       
\usepackage{url}            
\usepackage{booktabs}       
\usepackage{graphicx}       
\usepackage{wrapfig}        
\usepackage{subcaption}     
\usepackage{amsfonts}       
\usepackage{nicefrac}       
\usepackage{microtype}      
\usepackage{xcolor}         
\usepackage{colortbl}       
\usepackage{arydshln}  
\usepackage{multirow}

\definecolor{lightred}{rgb}{1.0, 0.9, 0.9}
\definecolor{badred}{HTML}{FDDEDE}
\definecolor{goodgreen}{HTML}{E0F0E0}
\definecolor{headblue}{HTML}{E8EEF8}

\usepackage{enumitem}
\usepackage{array}
\usepackage{multirow}
\usepackage[many]{tcolorbox}

\usepackage{amsthm}

\usepackage{algorithm}
\usepackage{algpseudocode}
\usepackage{float}
\usepackage{xspace}
\usepackage{pgfplots}
\pgfplotsset{compat=1.18}
\usepackage{cleveref}  
\definecolor{mygray}{gray}{0.95}
\definecolor{myyellow}{RGB}{255,245,200}
\definecolor{myblue}{RGB}{230,240,255}
\definecolor{deltared}{RGB}{180,0,0}
\definecolor{deltagreen}{RGB}{0,130,0}
\tcbset{
  aibox/.style={
    width=\linewidth,
    top=8pt,
    bottom=4pt,
    colback=blue!6!white,
    colframe=black,
    colbacktitle=black,
    fonttitle=\bfseries,
    enhanced,
  }
}
\definecolor{citeblue}{HTML}{4A90E2}   

\hypersetup{
    colorlinks=true,
    citecolor=citeblue,
    linkcolor=citeblue,
    urlcolor=citeblue,
    filecolor=citeblue
}

\newcolumntype{C}[1]{>{\centering\let\newline\\\arraybackslash\hspace{0pt}}m{#1}}
\newtcolorbox{AIbox}[2][]{aibox,title=#2,#1}

\tcbset{
  casebox/.style={
    width=\linewidth,
    top=8pt,
    bottom=4pt,
    colback=cyan!8!white,
    colframe=blue!50!black,
    colbacktitle=blue!50!black,
    fonttitle=\bfseries,
    enhanced,
  }
}
\newtcolorbox{CaseStudyBox}[2][]{casebox,title=#2,#1}

\tcbset{
  failcasebox/.style={
    width=\linewidth,
    top=8pt,
    bottom=4pt,
    colback=red!6!white,
    colframe=red!60!black,
    colbacktitle=red!60!black,
    fonttitle=\bfseries,
    enhanced,
  }
}
\newtcolorbox{FailCaseBox}[2][]{failcasebox,title=#2,#1}

\tcbuselibrary{listings,breakable}
\newtcblisting{PromptSheet}[1]{%
  breakable, enhanced, sharp corners=south,
  colback=gray!3!white, colframe=gray!50!black,
  coltitle=white, colbacktitle=gray!50!black,
  fonttitle=\bfseries\footnotesize, boxrule=0.5pt,
  left=4pt, right=4pt, top=2pt, bottom=2pt,
  title={#1}, listing only,
  listing options={basicstyle=\ttfamily\scriptsize, breaklines=true,
    breakindent=0pt, breakautoindent=false, columns=fullflexible,
    keepspaces=true, showstringspaces=false}
}

\newcommand{\Crafter}{\textsc{Crafter}\xspace}
\newcommand{\Editor}{\textsc{CraftEditor}\xspace}
\newcommand{\CrafterBench}{\textsc{CraftBench}\xspace}

\newcommand{\PaperBananaBench}{PaperBanana-Bench\xspace}

\newcommand{\AutoFigureEdit}{AutoFigure-Edit\xspace}
\newcommand{\EditBanana}{Edit-Banana\xspace}

\title{\Crafter: A Multi-Agent Harness for Editable Scientific Figure Generation from Diverse Inputs}

\author{
  Haozhe Zhao$^{1*}$ \quad Shuzheng Si$^{2*}$ \quad Zhenhailong Wang$^{1}$ \quad Zheng Wang$^{1}$ \quad Liang Chen$^{3}$ \\
  \textbf{Xiaotong Li$^{3}$ \quad Zhixiang Liang$^{1}$ \quad Maosong Sun$^{2}$ \quad Minjia Zhang$^{1}$} \\[0.5em]
  $^{1}$University of Illinois at Urbana-Champaign \quad $^{2}$Tsinghua University \quad $^{3}$Peking University \\
  \texttt{haozhez6@illinois.edu} \\[0.3em]
  {\small $^{*}$Equal contribution }
}

\begin{document}

\maketitle

\begin{abstract}
Scientific figures are among the most effective means of communicating complex research ideas, yet producing publication-quality illustrations remains one of the most labor-intensive parts of paper preparation. Existing automated systems each target a single figure type under text-only input, leaving the diversity of types and conditions researchers actually use unaddressed; their raster outputs further cannot be locally revised. Because scientific figures are structured compositions of discrete semantic components, the localized errors generators produce on such layouts demand not a stronger backbone but a \emph{harness}. We instantiate this harness in two complementary systems: \Crafter, a multi-agent harness for figure generation that generalizes across figure types and input conditions without architectural changes, and \Editor, which applies the same pattern to convert raster outputs into editable SVGs. Moreover, we introduce \CrafterBench, a benchmark spanning three figure types and
four input conditions with human quality annotation. Experiments show
that \Crafter\ substantially outperforms both standalone generators and
the agentic baseline on \PaperBananaBench\ and \CrafterBench,
with ablations confirming each component's independent contribution;
\Editor\ faithfully converts outputs into editable SVGs that surpass all baselines. Our code and benchmark are available at \url{https://github.com/HaozheZhao/Crafter}.
\end{abstract}

\section{Introduction}
\label{sec:intro}

Text-to-image generation has advanced rapidly, with recent models producing photorealistic and design-quality images across creative, medical, and scientific domains~\citep{qwen2025image,glm2025image,labs2024flux}. One area where this progress has yet to translate into practical tools is scientific illustration, where producing publication-quality figures remains one of the most labor-intensive parts of paper preparation. Recent work has begun to tackle this from two directions: agentic pipelines that pair planning agents with powerful image generators to produce visually polished figures from text~\citep{zhu2026paperbanana,zhu2026autofigure,sun2025p2p,guo2026paper2sysarch,kukreja2026cage,yang2026omnidiagram}, and code-generation methods that synthesize editable diagrams in TikZ or similar formats~\citep{belouadi2024automatikz,zala2024diagrammergpt,greisinger2026tikzilla,PPTAgent}. While encouraging, these approaches fall short of real-world demands in
two fundamental respects.

First, existing systems are narrow in scope. In practice, researchers produce figures across a spectrum of types, from academic diagrams to posters and infographics, and rarely begin from text alone; instead, they iterate from rough sketches, partial layouts, or reference viusal elements or icons. Current methods, by contrast, focus predominantly on text-to-image generation~\citep{zhu2026paperbanana,zhu2026autofigure,guo2026paper2sysarch,sun2025p2p}, leaving diversity of figure types and input conditions entirely unaddressed. Existing evaluations reflect same narrow scope, covering only text-to-image generation of methodology figures~\citep{zhu2026paperbanana} with no mechanism to assess whether a system generalizes across figure types or preserves a user's conditioning input.
Second, output images are not practically editable. Raster-based generators produce static images that cannot be locally revised, which is problematic when researchers need to adjust or revise individual labels, swap color schemes, or rearrange components. Code-generation methods yield editable output but lack the visual richness of icons and stylized layouts; recent raster-to-vector attempts~\citep{lin2026autofigureedit,editbanana2026} remain limited by unreliable element extraction and fragile composition. A complete scientific figure pipeline must therefore extend beyond generation to produce structurally editable output.

Addressing the generation challenge requires more than a more powerful backbone.
Scientific figures, unlike natural images, are structured
compositions of discrete semantic components: labeled boxes, directional arrows, icons, and annotations, each carrying specific meaning within precise spatial relationships.
Modern generators exhibit high output variance on such structured layouts, producing localized errors such as garbled labels and misaligned connectors that prompt rephrasing alone cannot fix. Naive retry is ineffective because each attempt produces a different constellation of failures, and accumulating free-text corrections across iterations introduces contradictions that further degrade quality. The same pattern holds for raster-to-vector conversion, where imprecise extraction and fragile composition persist across one-shot attempts. What is needed in both settings is not a better generator but a \emph{harness}~\citep{young2025harness,pan2026nlaharnesses,bui2026opendev, si2026contextskillslanguagemodels}: an orchestration layer that wraps an existing engine with an evolving structured specification as its memory, enabling targeted correction of individual failure points and closed-loop verification against the original intent.

\begin{figure}[t]
\centering
\includegraphics[width=\linewidth]{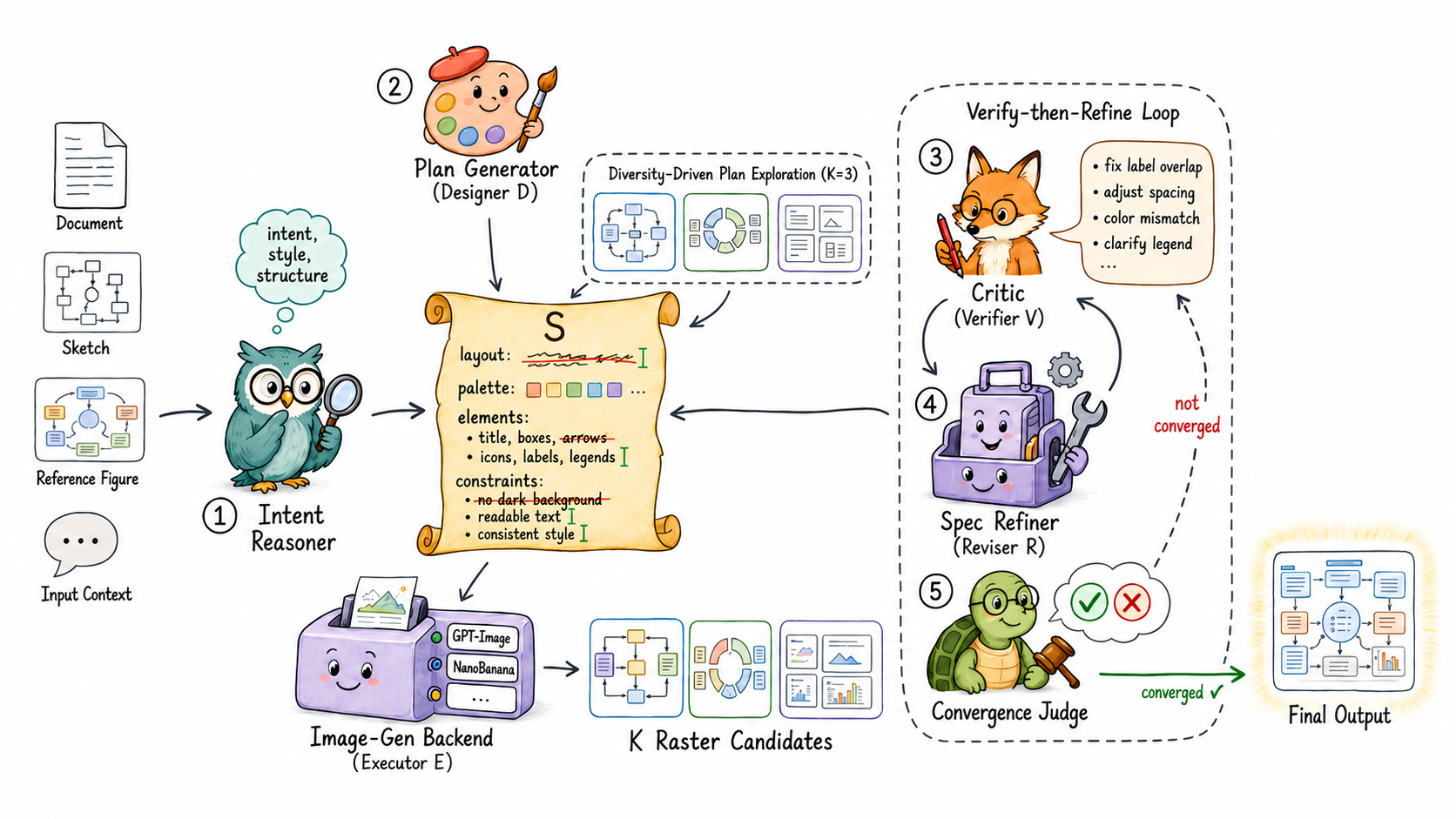}
\caption{\Crafter\ architecture. Given context and Docs, the intent reasoner seeds $\mathcal{S}_0$. The plan generator $\mathcal{D}$ proposes $K$ candidate plans; Image-Gen backend $\mathcal{E}$ renders each plan; the critic $\mathcal{V}$ emits directive diagnostics; the specification refiner $\mathcal{R}$ writes typed edits into $\mathcal{S}$; and the convergence judge routes each round to accept, refine, or revert to Final output. (Figure is generated by \Crafter.)}
\label{fig:crafter-architecture}
\end{figure}

We instantiate this harness in two complementary systems. \textbf{\Crafter} is a multi-agent harness for scientific figure generation in which cooperating agents, an intent reasoner, a plan generator, a critic, a specification refiner, and a convergence judge, share an evolving figure specification as the pipeline's structured memory, while an image-generation backend handles all rendering. 
Three mechanisms underpin the design: \emph{diversity-driven plan
exploration} generates multiple candidate framings in parallel; a \emph{structured corrective layer} accumulates critique-driven typed edits into the shared specification, preventing the prompt contradictions that plague free-text revision; and a \emph{verify-then-refine loop} in which a directive critic issues targeted corrections rather than scalar scores.
Because all task-specific behavior resides in agent prompts, the same architecture generalizes across figure types and input conditions without structural change.
\textbf{\Editor} applies the same harness pattern to convert raster figures into editable SVGs through three sequential phases. 
An extraction phase strips away text overlays and visual clutter to obtain clean graphical
assets from the original layout; a processing phase captions each asset and classifies it as vector or raster; and a composition phase assembles these assets into an SVG skeleton and iteratively refines the result via a hybrid critic. 

To evaluate across this broader scope, we introduce \textbf{\CrafterBench}, a $279$-sample benchmark spanning three figure types and four input conditions, curated from published papers across eighteen research areas, award-tier conference posters, and research blogs through a multi-stage pipeline with human quality annotation. Following previous work~\citep{zhu2026paperbanana}, we also adopt an evaluation protocol that assesses output quality against real images using VLMs as judges. Experiments show that \Crafter\ substantially outperforms both standalone generators and the strongest agentic baseline on \PaperBananaBench\ and \CrafterBench\ under controlled comparison. Ablations validate each mechanism, with removal of any single component causing a $5.04$ to $8.90$~point drop. \Editor\ converts generated outputs into editable SVGs, outperforming all baselines on a three-VLM ensemble evaluation, making our method a pioneering step toward the full generation-to-editing workflow.

Our contributions are summarized as follows:
\begin{itemize}[leftmargin=1.2em, itemsep=0pt, topsep=0pt, parsep=0pt]
  \item A unified harness framework instantiated as \textbf{\Crafter}
  for cross-type, cross-condition scientific figure generation and
  \textbf{\Editor} for raster-to-SVG conversion, together forming the
  first end-to-end generation-to-editing pipeline for scientific figures.
  \item \textbf{\CrafterBench}, a benchmark spanning three
  figure types and four input conditions with human quality annotation,
  paired with a VLM-based evaluation protocol for conditional figure
  assessment.
  \item State-of-the-art results on both benchmarks, with detailed
  ablations.
\end{itemize}

\section{Related Work}
\label{sec:related}
 
\noindent\textbf{Scientific figure creation.}\
Automated scientific figure creation falls into two families. Code-generation methods synthesize editable diagrams in code like TikZ from text descriptions~\citep{belouadi2024automatikz,zala2024diagrammergpt,greisinger2026tikzilla}, but are restricted to schematic diagrams and lack the visual richness of icons and stylized layouts. Agentic pipelines pair LLM agents with image generators to produce high-quality raster figures for methodology plots~\citep{zhu2026paperbanana,zhu2026autofigure}, posters~\citep{sun2025p2p}, architecture diagrams~\citep{guo2026paper2sysarch}, and educational illustrations~\citep{kukreja2026cage,yang2026omnidiagram}, yet their raster outputs cannot be easily revised. Recent efforts to bridge raster output and editability remain preliminary: AutoFigure-Edit~\citep{lin2026autofigureedit} detects elements and emits an SVG in a single LLM call, and Edit-Banana~\citep{editbanana2026} converts segmentation and OCR outputs into DrawIO cells. Across both families, a shared limitation persists: each system targets a single figure type, accepts only text input, and ignores the diversity of conditions in real research workflows. Work on agent orchestration~\citep{young2025harness,pan2026nlaharnesses} and iterative self-correction~\citep{madaan2023selfrefine} has shown that the harness layer matters as much as the underlying model, but this principle remains unexplored for scientific figures.

\noindent\textbf{Benchmarks and evaluation.}\
Evaluation method for scientific figure generation remains as narrow as systems it measures. PaperBanana-Bench~\citep{zhu2026paperbanana} and Paper2SysArch~\citep{guo2026paper2sysarch} evaluate only text-to-image generation of academic diagrams. SridBench~\citep{chang2025sridbenchbenchmarkscientificresearch} covers thirteen fields but remains limited to text-to-image generation without conditional inputs. IGenBench~\citep{tang2026igenbench} targets text-to-infographic reliability with a decomposed verification framework but covers only infographics. SciFlow-Bench~\citep{zhang2026sciflowbench} inverse-parses generated diagrams into structured graphs to measure structural recoverability, but is limited to flowchart diagrams. None of these benchmarks tests cross-type, cross-condition
generalization. \CrafterBench\ fills this gap with coverage of
three figure types and four input conditions.

\section{Method}
\label{sec:method}

As analyzed in Section~\ref{sec:intro}, generating scientific figures
reliably faces three technical difficulties: high output variance on
complex structured layouts, prompt degradation from accumulated free-text
corrections, and the absence of structured corrective feedback. These
difficulties call for a harness layer that orchestrates planning,
verification, and revision around the generator rather than improving the
generator itself (\S\ref{sec:method-harness}). To make this harness
effective, we equip it with three targeted mechanisms:
diversity-driven plan exploration (\S\ref{sec:method-diversity}), a
structured corrective layer (\S\ref{sec:method-corrective}), and
verify-then-refine iteration with a directive critic
(\S\ref{sec:method-verify}). We instantiate these mechanisms in
\Crafter, a multi-agent harness for figure generation
(\S\ref{sec:method-crafter}). To further make all generated figures
editable, \Editor\ reuses the same harness pattern for raster-to-vector
conversion (\S\ref{sec:method-editor}).

\subsection{The Harness Abstraction}
\label{sec:method-harness}

A \emph{harness} is an orchestration layer that wraps a executor with planning, verification, and structured revision, detecting and correcting executor's failure modes without modifying executor itself. In our setting the executor is an image generator (for \Crafter) or a code generator (for \Editor).
We formalize harness as a four-role loop over a shared \emph{evolving specification} $\mathcal{S}$, a structured record that accumulates current plan, revision history, and prior diagnostics (Figure~\ref{fig:crafter-architecture}). At each round $t$:
\begin{align}
  p_t &= \mathcal{D}(\text{input},\;\mathcal{S}_{t-1}),
  &\quad a_t &= \mathcal{E}(p_t), \label{eq:harness-de}\\
  d_t &= \mathcal{V}(a_t,\;\text{input},\;\mathcal{S}_{t-1}),
  &\quad \mathcal{S}_t &= \mathcal{R}(d_t,\;\mathcal{S}_{t-1}),
  \label{eq:harness-vr}
\end{align}
where designer $\mathcal{D}$ produces an actionable plan $p_t$, executor $\mathcal{E}$ renders it into an artifact $a_t$, verifier $\mathcal{V}$ emits a \emph{directive diagnostic} $d_t$ (per-dimension scores, identified defects, and suggested corrections, as opposed to a scalar quality score), and reviser $\mathcal{R}$ applies \emph{typed edits} to $\mathcal{S}_{t-1}$, each a structured operation (adding a layout constraint, banning an artifact category, resizing a named element) that modifies the specification in place rather than appending free text to the prompt. The loop terminates when $\mathcal{V}$ accepts $a_t$ or a round budget $T$ is reached, returning $a^{*}\!=\!\arg\max_\tau\;\mathrm{score}(d_\tau)$.

Two properties make this loop effective for scientific figures:
$\mathcal{E}$ is pluggable, so all task-specific behavior resides in the
prompts of $\mathcal{D}$, $\mathcal{V}$, and $\mathcal{R}$; and
$\mathcal{R}$ writes typed edits to a shared record rather than
free-text additions to the prompt, keeping the specification internally
consistent across rounds. Table~\ref{tab:harness-roles} summarizes how
\Crafter\ and \Editor\ instantiate each role.

\begin{table}[t]
\centering
\small
\caption{Harness role assignments for \Crafter\ and \Editor.}
\label{tab:harness-roles}
\begin{tabular}{@{}lll@{}}
\toprule
Role & \Crafter & \Editor \\
\midrule
$\mathcal{D}$ Designer  & Plan generator        & SVG skeleton generator \\
$\mathcal{E}$ Executor  & Image-generation backend & Element-injection code \\
$\mathcal{V}$ Verifier  & Multi-dim.\ critic    & Hybrid critic \\
$\mathcal{R}$ Reviser   & Specification refiner & SVG editor \\
\bottomrule
\end{tabular}
\end{table}

\subsection{\Crafter: Harness for Figure Generation}
\label{sec:method-crafter}


\Crafter\ works as a harness for scientific figure generation. Given a context $\mathbf{c}$ (e.g., papers, reference images, or sketches) and an instruction $\mathbf{q}$, it produces a publication-quality raster figure $a^*$ together with the final specification $\mathcal{S}_T$. 
As established in \S\ref{sec:method-harness}, the same pipeline
generalizes across diverse figure types and input conditions through
prompt-level adaptation alone.

Five cooperating agents implement the four harness roles (Figure~\ref{fig:crafter-architecture}; Table~\ref{tab:harness-roles}). An \emph{intent reasoner} analyzes $(\mathbf{c}, \mathbf{q})$ and infers the figure's communicative role and required visual elements, seeding the initial specification $\mathcal{S}_0$. The plan generator $\mathcal{D}$ reads $\mathcal{S}_0$ and proposes candidate visual plans; the image-generation backend $\mathcal{E}$ renders each plan into a raster; the critic $\mathcal{V}$ evaluates every candidate against $\mathcal{S}$ and the original input $(\mathbf{c}, \mathbf{q})$; and the specification refiner $\mathcal{R}$ writes typed edits back into $\mathcal{S}$. A convergence judge governs the loop at each round, deciding whether to accept, continue refining, or revert to $a^*$. Three mechanisms, detailed below, address the three failure modes identified in Section~\ref{sec:intro}; full agent prompts are provided in Appendix~\ref{app:crafter-impl}.

\subsubsection{Diversity-Driven Plan Exploration}
\label{sec:method-diversity}

Modern image generators exhibit high inter-sample variance on complex, structured figures: qualitatively different layouts and compositions emerge across random seeds for a fixed prompt. A single-draw pipeline cannot recover from a structurally unsuitable sample, and hardcoding a fixed style restricts the generation space unnecessarily. \Crafter\ treats this variance as a search problem: $\mathcal{D}$ reads $\mathcal{S}_0$ and proposes $K$ intent-conditioned candidate plans, each specifying a distinct visual framing (e.g., banner layout or multi-column grid). $\mathcal{E}$ renders all $K$ plans in parallel, and the convergence judge selects the best candidate $a^{(1)} \!=\! \arg\max_{k} \mathrm{score}(\mathcal{V}(a_k))$ as the starting point for subsequent refinement. $K$ is set adaptively based on input constraints. Unlike additional refinement rounds, plan-level branching can escape a fundamentally unsuitable compositional choice before any rendering budget is spent on refining it.

\subsubsection{Structured Corrective Layer}
\label{sec:method-corrective}

Iterative repair via free-text revision instructions degrades rapidly: successive natural-language addenda introduce conflicting directives (e.g., ``enlarge the title'' followed by ``reduce white space''), the generator absorbs the contradictions silently, and faithfulness deteriorates without any single round appearing anomalous. The structured corrective layer replaces free-text accumulation with \emph{typed edits} on $\mathcal{S}$: at each round $t$, $\mathcal{R}$ converts the diagnostic $d_t$ into a set of structured operations $\{e_i\} = \mathcal{R}(d_t, \mathcal{S}_{t-1})$, where each $e_i$ modifies $\mathcal{S}$ in place (adding a layout constraint, banning an artifact category, resizing a named element). The next round's prompt is assembled from this coherent record $\mathcal{S}_t$ rather than from a growing stack of amendments, keeping the specification internally consistent across rounds. 

\subsubsection{Verify-then-Refine with a Directive Critic}
\label{sec:method-verify}

Even with a well-chosen plan and an accumulating specification, first-generation outputs typically contain localized errors, such as missing components or duplicated regions. This mechanism comprises two components, a \emph{directive critic} that diagnoses errors and a \emph{verify-then-refine loop} that applies the corrections iteratively, each validated independently.

\textbf{Directive critic.}\quad A scalar score (e.g., ``5/10'') provides no actionable target for the next round. $\mathcal{V}$ instead emits a directive diagnostic $d_t = \mathcal{V}(a_t,\,\mathbf{c},\,\mathbf{q},\,\mathcal{S}_{t-1})$ containing per-dimension scores along six axes, identified defects, suggested corrections, and a revised figure description. $\mathcal{R}$ converts $d_t$ into edits on $\mathcal{S}$ (\S\ref{sec:method-corrective}), and the prompt builder injects the corrections as fix-guidance into the next round.

\textbf{Refinement loop.}\quad An \emph{early-exit gate} bypasses the loop when the first-round output already satisfies acceptance thresholds on critical dimensions. Otherwise the loop runs for up to $T{=}3$ rounds, with a \emph{best-so-far checkpoint} that reverts to $a^*$ whenever the current round regresses, since language-model-driven iterative editing is empirically non-monotonic~\citep{madaan2023selfrefine}.

\subsection{\Editor: Harness for Raster-to-Vector Conversion}
\label{sec:method-editor}


Raster figures do not support the element-level edits that research workflows demand (e.g., swapping icons, or completing a partial diagram). \Editor\ converts a raster figure $a^*$, whether produced by \Crafter\ or obtained externally, into a coordinate-faithful editable SVG $\mathbf{v}$ by instantiating the same harness loop (Eqs.~\ref{eq:harness-de}--\ref{eq:harness-vr}) on vector composition rather than pixel synthesis.

\begin{figure}[t]
\centering
\includegraphics[width=0.96\linewidth]{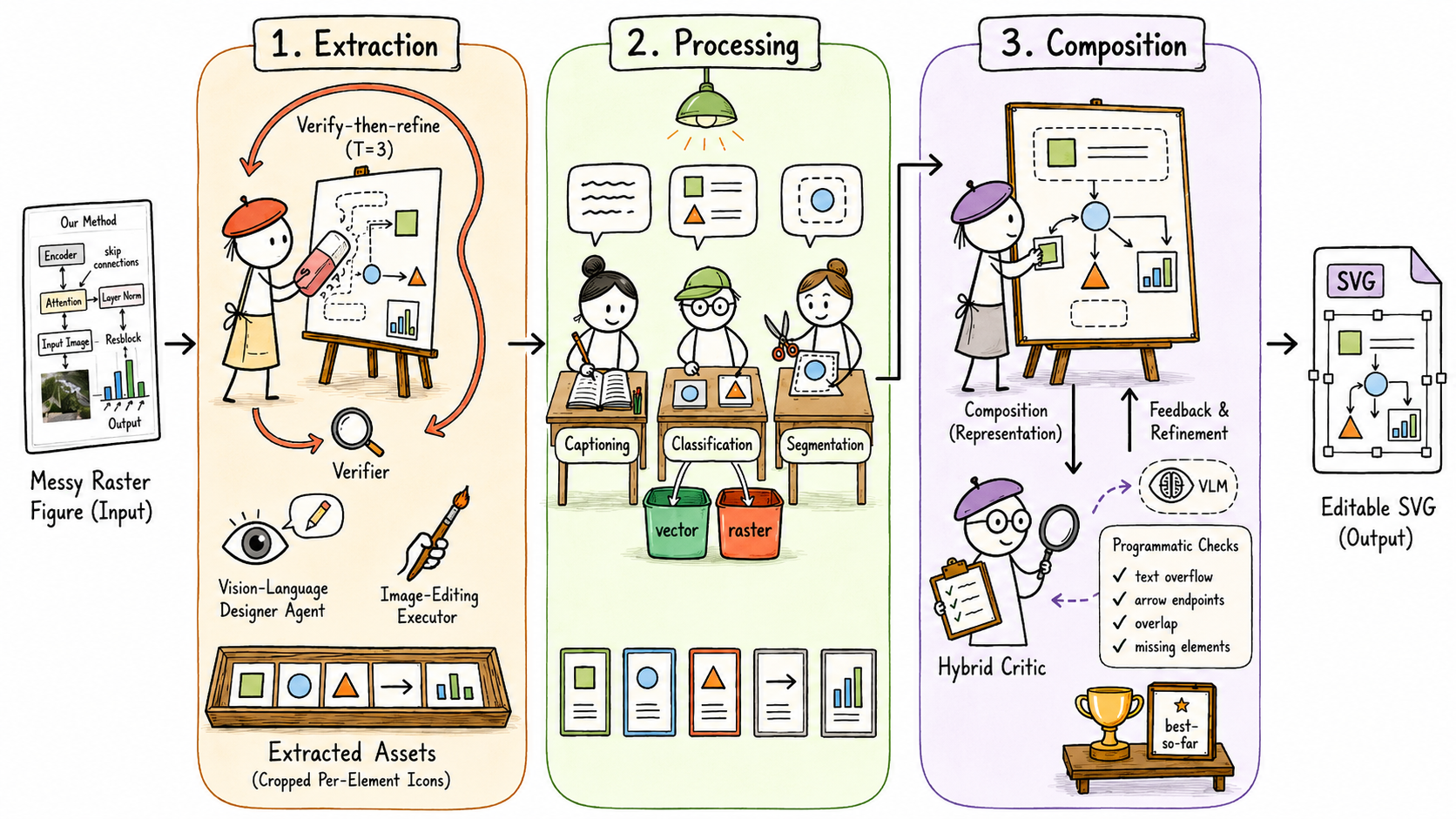}
\caption{\Editor\ architecture. Three phases convert a raster $a^*$ into an editable SVG $\mathbf{v}$. \emph{Extraction}: a VLM designer $\mathcal{D}$ authors a keep/delete plan, an image editor $\mathcal{E}$ executes it, and $\mathcal{V}$ verifies the cleaned canvas (\S\ref{sec:method-editor-extraction}). \emph{Processing}: each element is captioned, grounded, and classified. \emph{Composition}: $\mathcal{D}$ generates SVG skeletons, $\mathcal{E}$ injects assets, and a hybrid critic $\mathcal{V}$ (VLM $+$ programmatic checkers) drives iterative refinement (\S\ref{sec:method-editor-composition}). (The Figure was generated by \Crafter.)}
\label{fig:editable-output-pipeline}
\end{figure}

Three phases organize the harness (Figure~\ref{fig:editable-output-pipeline}; Table~\ref{tab:harness-roles}). An \emph{extraction phase} strips visual clutter from $a^*$ and isolates per-element assets. A \emph{processing phase} captions, grounds, and classifies each element (Appendix~\ref{app:editable-ablations}). A \emph{composition phase} assembles the assets into a final SVG and refines it through a critic-driven loop. The extraction and composition phases each instantiate the four-role harness: $\mathcal{D}$ authors a plan (a keep/delete specification for extraction, an SVG skeleton for composition), $\mathcal{E}$ executes it, $\mathcal{V}$ inspects the result, and $\mathcal{R}$ revises the plan based on $d_t$.


\subsubsection{Extraction: Instruction-Driven Canvas Cleaning}
\label{sec:method-editor-extraction}

Scientific figures, particularly conference posters containing 25 to 50
visual assets, exhibit overlapping elements, text, and heterogeneous
backgrounds that defeat off-the-shelf segmentation, which struggles to
produce reliable boundaries and distinguish semantically relevant
components on such cluttered layouts. \Editor\ replaces segmentation with
an instruction-driven extraction loop. $\mathcal{D}$ (a vision-language
agent) inspects $a^*$ and authors a per-figure keep/delete plan $p_t$
specifying which elements to preserve and which to remove. $\mathcal{E}$
(an instructable image editor) executes the plan at the pixel level,
producing a cleaned canvas $a_t$. $\mathcal{V}$ inspects $a_t$ and
either accepts it or returns a diagnostic $d_t$ that triggers another
round, for at most $T{=}3$ iterations. Per-element assets are then
cropped from the clean canvas, with a hallucination filter discarding
blank, mismatched, or text-only extractions before they reach the
composition phase.

\subsubsection{Composition: Iterative SVG Assembly}
\label{sec:method-editor-composition}

A single language-model call to produce an SVG from the element inventory routinely generates layouts whose grid topology, arrow endpoints, or text labels disagree with the input raster. The composition phase replaces this one-shot call with the full harness loop.

$\mathcal{D}$ generates two candidate SVG skeletons at different decoding temperatures; the convergence judge selects the better candidate via a rapid visual comparison. $\mathcal{E}$ splices the extracted assets into the placeholders of the selected skeleton. $\mathcal{V}$ then evaluates the rendered SVG against $a^*$ via a \emph{hybrid critic} that combines two complementary channels: a vision-language model assessing global layout fidelity and semantic correspondence, and programmatic checkers auditing structural properties (text overflow, arrow-endpoint accuracy, element overlap, missing components) that vision-language evaluation alone tends to miss. $\mathcal{R}$ modifies the SVG source in response to $d_t$. The loop runs for up to $T{=}4$ rounds, with best-so-far reversion to $a^*$ guarding against non-monotonic regressions. As in \Crafter, all task-specific behavior resides in prompts (Appendix~\ref{app:editable-ablations}).

\begin{figure}[h]
\centering
\includegraphics[width=\linewidth]{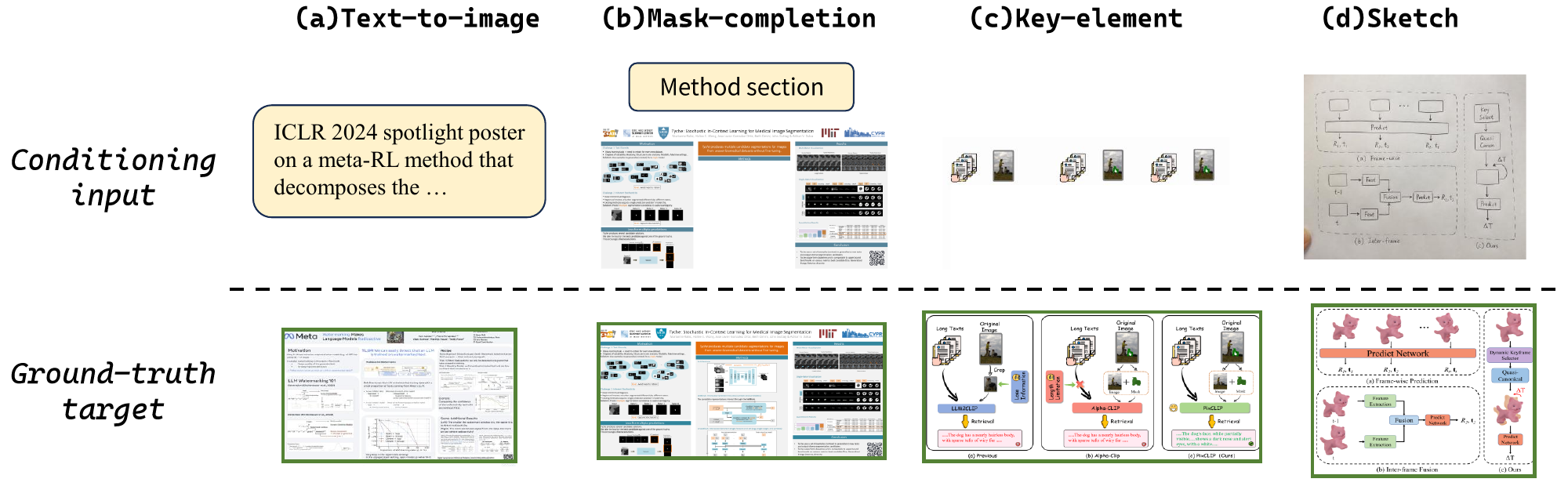}
\caption{Representative \CrafterBench\ samples.
Each column shows one task.}
\label{fig:craftbench-examples}
\end{figure}

\section{\CrafterBench}
\label{sec:bench}


\CrafterBench\ evaluates scientific figure generation across three
figure types and four input conditions (text-to-image and three
reference-conditioned tasks: mask-completion, key-element composition,
and sketch-conditioned generation), totaling $279$ curated samples. Figure~\ref{fig:craftbench-examples} shows representative samples
per task, illustrating the conditioning input and the ground-truth target.

\subsection{Data Construction}
\label{sec:bench-data}

Samples are built by a three-stage pipeline (full details in Appendix~\ref{app:bench-details}). \emph{Collection} draws academic figures from arXiv preprints across $18$ subject areas, posters from award-tier conference papers, and infographics from long-form research blogs \citep{si-etal-2025-gateau}. \emph{Filtering} applies vision-language content classification, complexity scoring, and claim-alignment verification, leaving $553$ candidates that a human curation reduces to the final $279$ samples balanced across tasks and styles. \emph{Annotation} pairs each text-to-image sample with its caption and source paper-text, and constructs a conditioning input for the three reference-conditioned tasks. Every reference-conditioned sample is reviewed by three graduate-level annotators through a per-task interface and accepted only on unanimous agreement, with disagreements triggering revision until consensus.

\subsection{Statistics}
\label{sec:bench-stats}

\begin{wrapfigure}{r}{0.40\linewidth}
\vspace{-6.0em}
\centering
\includegraphics[width=\linewidth]{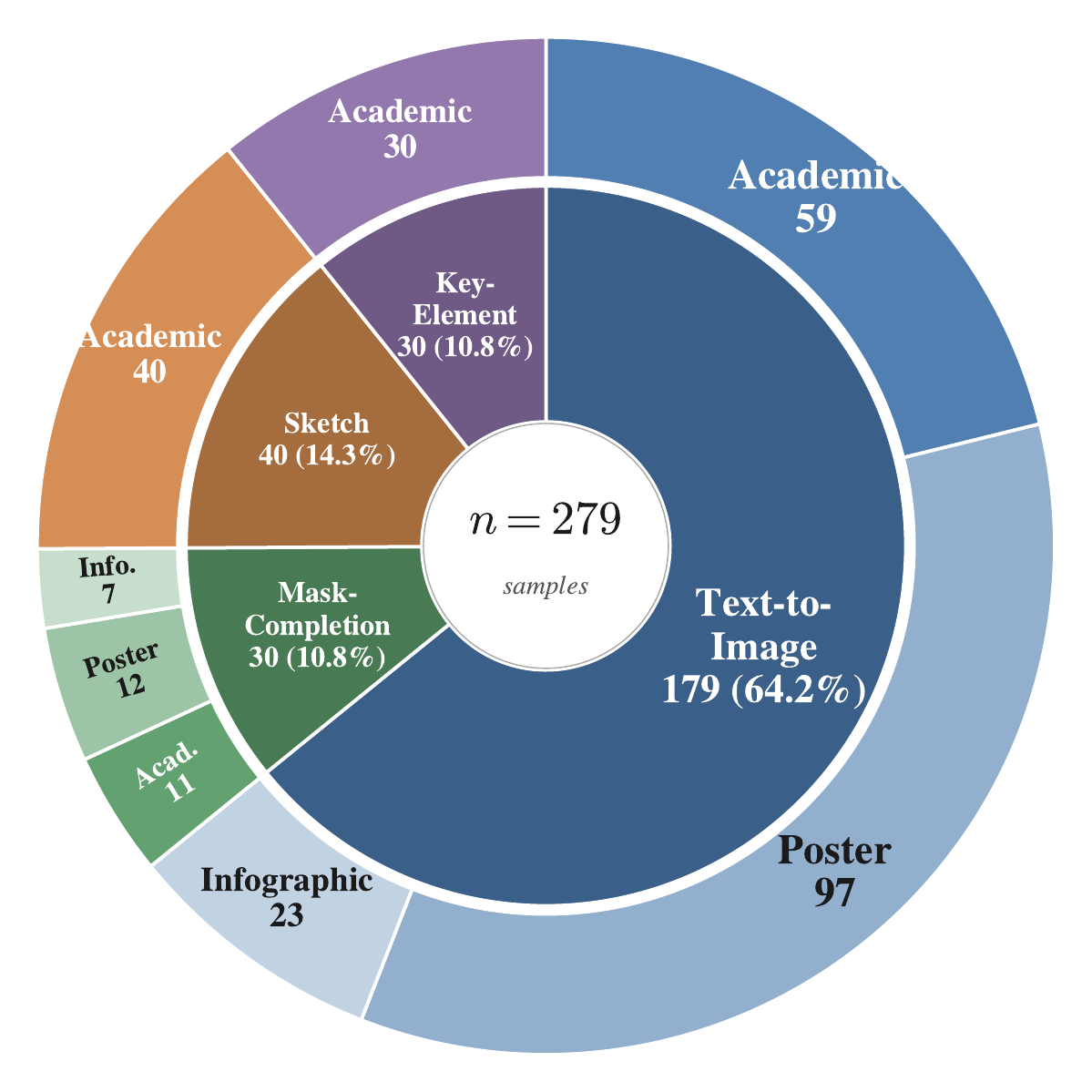}
\vspace{-2.0em}
\caption{\CrafterBench\ distribution. Inner: task types; outer: per-task style.}
\label{fig:craftbench-distribution}
\vspace{-4.0em}
\end{wrapfigure}

\CrafterBench\ contains $279$ samples spanning four tasks and three styles (Figure~\ref{fig:craftbench-distribution}). Text-to-image accounts for nearly two-thirds of the benchmark ($n{=}179$), with the three reference-conditioned tasks contributing mask-completion ($30$), sketch-conditioned ($40$), and key-element composition ($30$). Text-to-image and mask-completion span all three style families, while the sketch and key-element tasks are drawn entirely from academic figures. Academic figures form the largest share ($140$), complemented by $109$ posters and $30$ infographics.


\subsection{Evaluation Protocol}
\label{sec:bench-judge}

Our evaluation keeps the referenced VLM-as-judge philosophy of \citet{zhu2026paperbanana}, scoring each output against the human-drawn target and reporting a lenient win-rate, but redesigns the judge for the cross-type, cross-condition setting. A Gemini~3.5~Flash~\citep{gemini35flash} judge scores the candidate and the target \emph{independently}, one image at a time rather than side by side, which removes the position bias of pairwise comparison. Scoring uses a compact set of task- and content-type-specific aspects rated from $0$ to $10$. Text-to-image samples are rated on content faithfulness, readability, and a style-specific format aspect for academic, poster, or infographic figures. The three reference-conditioned tasks replace the format aspect with an input-fidelity aspect tailored to how each task uses its conditioning input. A weighted mean turns the per-aspect scores into one total per image, and the candidate's margin over the target yields a verdict $o_i \in \{\textit{Model},\,\textit{Tie},\,\textit{Human}\}$ under a calibrated tie band. The bench-level score averages the $\{100, 50, 0\}$ mapping of these verdicts, and on academic text-to-image inputs it reduces to a PaperBanana-style referenced judge. A blind human study on a random sample confirms that this metric tracks human preference (Appendix~\ref{app:human-eval}). Full prompts are in Appendix~\ref{app:eval-protocol}.

\section{Experiments}
\label{sec:experiments}

We evaluate \Crafter\ on two benchmarks: \PaperBananaBench~\citep{zhu2026paperbanana}, which covers text-to-image generation of academic figures, and our proposed \CrafterBench, which extends coverage to three figure types and four input conditions. Both benchmarks are scored by referenced VLM-as-judge protocols that report a lenient win-rate against the human-drawn target (\S\ref{sec:bench-judge}). We compare against standalone generators and agentic frameworks; to isolate the effect of orchestration design, all agentic methods share the same image-generation backbone (Nano~Banana~2) and vision-language model (Gemini~3.1~Pro~\citep{gemini31pro}). Full configuration details are in Appendix~\ref{app:experiment_setup}.

\begin{table*}[t]
\centering
\caption{Results (\%) on \PaperBananaBench\ and \CrafterBench. \textbf{Bold} marks column-best; $\Delta$ is the gap between \Crafter\ and its standalone generator. $^{*}$On \CrafterBench, GPT-Image-2 returned valid outputs for only 260 of 279 inputs, likely due to instability and content-safety refusals.}
\label{tab:main-results}
\scriptsize
\setlength{\tabcolsep}{3pt}
\resizebox{\textwidth}{!}{
\begin{tabular}{lrrrrrrrrrr}
\toprule
\multicolumn{1}{l}{\multirow{2}{*}{\textbf{Method}}}
  & \multicolumn{5}{c}{\textbf{PaperBanana-Bench}}
  & \multicolumn{5}{c}{\textbf{CraftBench}} \\
\cmidrule(lr){2-6} \cmidrule(lr){7-11}
& \textbf{Faith.} & \textbf{Conc.} & \textbf{Read.} & \textbf{Aesth.} & \textbf{Overall}
& \textbf{T2I} & \textbf{Mask} & \textbf{Sketch} & \textbf{KeyEl.} & \textbf{Overall} \\
\midrule
\multicolumn{11}{l}{\emph{Standalone generators (open-source)}} \\
GLM-Image & 0.00 & 0.69 & 0.00 & 1.55 & 0.00 & 0.00 & 0.00 & 0.00 & 0.00 & 0.00 \\
Qwen-Image & 0.00 & 0.00 & 0.00 & 1.71 & 0.00 & 0.00 & 0.00 & 2.50 & 0.00 & 0.40 \\
\midrule
\multicolumn{11}{l}{\emph{Standalone generators (closed-source)}} \\
GPT-Image-2$^{*}$ & 8.42 & 3.97 & 1.72 & 40.72 & 1.37 & 5.30 & 18.30 & 45.00 & 36.70 & 15.80 \\
Nano Banana 2 & 15.07 & 11.99 & 26.88 & 47.95 & 11.13 & 8.40 & 30.00 & 61.20 & 23.30 & 19.90 \\
Nano Banana Pro & 25.43 & 36.13 & 37.16 & 50.52 & 22.43 & 17.30 & 21.70 & 46.20 & 21.70 & 22.40 \\
\midrule
\multicolumn{11}{l}{\emph{Agentic frameworks}} \\
AutoFigure (w/ Nano Banana 2) & 12.33 & 6.85 & 3.26 & 7.71 & 1.37 & 1.40 & 0.00 & 8.80 & 0.00 & 2.20 \\
PaperBanana (w/ Nano Banana 2) & 28.10 & 52.41 & 42.64 & 61.68 & 33.73 & 18.70 & 36.70 & 60.00 & 31.70 & 28.00 \\
PaperBanana (w/ Nano Banana Pro) & 31.34 & 52.74 & 45.03 & 61.30 & 35.96 & 26.80 & 23.30 & 45.00 & 26.70 & 29.00 \\
\rowcolor{blue!8} \textbf{\Crafter\ (w/ Nano Banana 2)} & \textbf{38.18} & 53.42 & 47.77 & 64.21 & \textbf{50.34} & 48.30 & \textbf{45.00} & 70.00 & \textbf{40.00} & 50.20 \\
\rowcolor{gray!8} \quad $\Delta$ vs.\ Nano Banana 2 & +23.11 & +41.43 & +20.89 & +16.26 & +39.21 & +39.90 & +15.00 & +8.80 & +16.70 & +30.30 \\
\rowcolor{blue!8} \textbf{\Crafter\ (w/ Nano Banana Pro)} & 37.72 & \textbf{55.31} & \textbf{48.46} & \textbf{65.29} & 50.00 & \textbf{52.50} & 41.70 & \textbf{73.80} & 33.30 & \textbf{52.30} \\
\rowcolor{gray!8} \quad $\Delta$ vs.\ Nano Banana Pro & +12.29 & +19.18 & +11.30 & +14.77 & +27.57 & +35.20 & +20.00 & +27.60 & +11.60 & +29.90 \\
\bottomrule
\end{tabular}
}
\end{table*}

\subsection{Main Results}
\label{sec:exp-headline}

Table~\ref{tab:main-results} presents the full comparison. \Crafter\ achieves the highest overall score on both benchmarks regardless of its backbone, leading the strongest agentic baseline under controlled comparison by $16.61$~point on \PaperBananaBench\ and $22.20$~point on \CrafterBench, and improving over its standalone generator on every quality dimension and every task. Among all methods, only \Crafter\ improves over its backbone uniformly, across every dimension and every task on both benchmarks. PaperBanana also improves over its backbone overall, but its gain shrinks sharply on the broader benchmark, from $22.60$~point on \PaperBananaBench\ to $8.10$~point on \CrafterBench, and it slips below its backbone on the sketch task. This is the generalization failure identified in Section~\ref{sec:intro}, where a pipeline optimized for a single figure type and input condition transfers poorly to broader settings. AutoFigure degrades on both benchmarks.

\begin{figure}[t]
\centering
\includegraphics[width=0.99\linewidth]{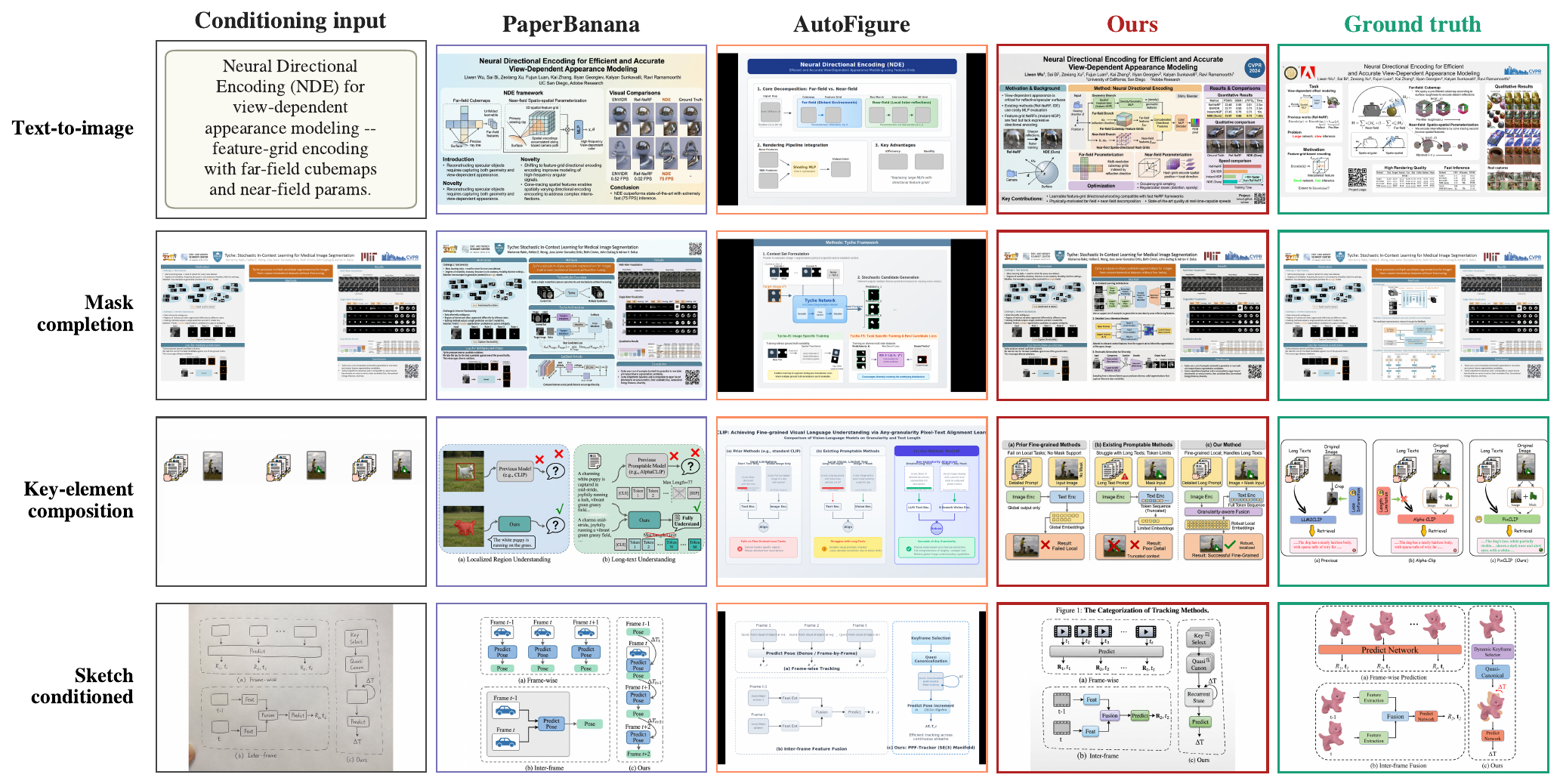}
\caption{Qualitative comparison across different input conditions.}
\label{fig:conditioning-qualitative}
\end{figure}

A per-task breakdown on \CrafterBench\ confirms that this advantage reflects broad generalization rather than strength on any single condition, as illustrated in Figure~\ref{fig:conditioning-qualitative}. Across its two backbones, \Crafter\ attains the best score in every column of both benchmarks, the four quality dimensions of \PaperBananaBench\ and all four tasks of \CrafterBench, indicating that the harness systematically strengthens the generation process rather than exploiting a narrow sweet spot. No baseline surpasses \Crafter\ in any column, and the strongest non-\Crafter\ results are scattered across different systems and conditions, so each baseline's strength is confined to specific inputs at the expense of general capability. The harness, by contrast, does not raise the generator's output ceiling but makes the pipeline more general and robust in handling diverse inputs and producing structurally sound layouts. The $\Delta$ rows in Table~\ref{tab:main-results} further show that this contribution is largely independent of the underlying executor: replacing Nano~Banana~2 with Nano~Banana~Pro shifts the overall score by only $0.34$~point on \PaperBananaBench\ and $2.10$~point on \CrafterBench, with neither backbone dominating uniformly, confirming that stronger future generators can be incorporated without modification. \Crafter\ is not uniformly successful, and Appendix~\ref{app:cases} and~\ref{app:failures} analyzes representative success and failure cases.


\subsection{Ablation and Analysis}
\label{sec:exp-analysis}



\begin{table}[t]
\centering
\caption{Mechanism ablation on \PaperBananaBench. Each row removes one mechanism from the full \Crafter\ harness. \textbf{Bold}: best per column; $\Delta$: overall gap vs.\ full \Crafter.}
\label{tab:ablation-pb}
\small
\setlength{\tabcolsep}{5pt}
\renewcommand{\arraystretch}{1.10}
\resizebox{0.8\linewidth}{!}{
\begin{tabular}{@{}l cccc cc@{}}
\toprule
\textbf{Configuration} & \textbf{Faith.} & \textbf{Conc.} & \textbf{Read.} & \textbf{Aesth.} & \textbf{Overall} & \textbf{$\Delta$} \\
\midrule
\rowcolor{blue!5}
\textbf{\Crafter} & \textbf{38.18} & \textbf{53.42} & \textbf{47.77} & \textbf{64.21} & \textbf{50.34} & \\
\addlinespace[3pt]
\quad \textit{w/o plan exploration}    & 28.42 & 51.20 & 38.30 & 60.10 & 41.78 & \textcolor[rgb]{0,0.5,0}{$-$8.56} \\
\quad \textit{w/o corrective layer}    & 27.18 & 49.66 & 36.86 & 58.45 & 41.44 & \textcolor[rgb]{0,0.5,0}{$-$8.90} \\
\quad \textit{w/o refinement loop}     & 30.07 & 51.97 & 41.55 & 61.80 & 44.86 & \textcolor[rgb]{0,0.5,0}{$-$5.48} \\
\quad \textit{w/o directive critic}    & 31.20 & 52.91 & 42.55 & 63.18 & 45.30 & \textcolor[rgb]{0,0.5,0}{$-$5.04} \\
\bottomrule
\end{tabular}
}
\end{table}

\subsubsection{Ablation Study}
\label{sec:exp-ablation-crafter}

We conduct ablation study on \PaperBananaBench\ to test the contribution of each mechanism independently, by removing one at a time from the full \Crafter\ pipeline. Results are shown in Table~\ref{tab:ablation-pb}. Every removal degrades the overall score, with drops ranging from $5.04$ to $8.90$~point.

Restricting \Crafter\ to a single candidate plan ($K{=}1$) incurs a $8.56$~point drop, as shown in \textit{w/o plan exploration}. Readability suffers the most among four quality dimensions, because a wrong framing decision early on, such as rendering a comparison grid as a block diagram, propagates through every subsequent refinement round with no opportunity to escape.
The \textit{w/o corrective layer} experiment shows that replacing typed edits with free-text revision instructions costs $8.90$~point overall. This result validates the core concern raised in Section~\ref{sec:intro}: when corrections accumulate as unstructured text, contradictions build silently across rounds and the generator's faithfulness erodes without any individual round appearing anomalous.
The \textit{verify-then-refine loop} and its \textit{directive critic} contribute $5.48$ and $5.04$~point respectively. Removing the loop confirms that iterative correction is essential for repairing localized errors that survive the first generation. The directive critic's independent contribution confirms that per-dimension diagnostics provide actionable targets that scalar scores cannot: without them, the reviser still iterates but lacks a specific failure to address.
We further analyze scaling behavior of $K$, $T$, and computational cost in Appendix~\ref{app:scaling} and~\ref{app:cost}.


\subsubsection{Editable-Output Quality}
\label{sec:exp-editable}

Raster outputs cannot be locally revised, so we evaluate \Editor's ability to convert \Crafter\ rasters into editable SVGs. We compare against Edit-Banana~\citep{editbanana2026} and AutoFigure-Edit~\citep{lin2026autofigureedit} on seven axes, scored by an ensemble of three VLM judges over a held-out subset of $80$ \Crafter\ outputs. Setup details are in Appendix~\ref{app:setup-editor} and Appendix~\ref{app:editable-ablations}.

\begin{table}[t]
\centering
\caption{Editable-output evaluation on $80$ \Crafter\ outputs. Scores ($0$--$10$): mean of three VLM judges. $\Delta$: drop vs.\ \Editor.}
\label{tab:editable_optimized}
\resizebox{0.85\linewidth}{!}{
    \small
    \setlength{\tabcolsep}{6pt}
    \renewcommand{\arraystretch}{1.1}
    \begin{tabular}{l r r r r r r r}
    \toprule
    \textbf{System} & \textbf{Pos.} & \textbf{Color} & \textbf{Text} & \textbf{Icon} & \textbf{Arrow} & \textbf{Style} & \textbf{Overall} \\
    \midrule
    \multicolumn{8}{l}{\textit{Raster-to-editable methods}} \\
    \EditBanana        & 4.21 & 4.93 & 2.86 & 4.97 & 4.18 & 4.32 & 3.69 \\
    \AutoFigureEdit    & 6.92 & 7.41 & 6.04 & 6.78 & 6.39 & 7.00 & 6.91 \\
    \midrule
    \rowcolor{blue!5} \textbf{\Editor} & \textbf{8.10} & \textbf{8.34} & \textbf{7.61} & \textbf{8.07} & \textbf{7.83} & \textbf{8.12} & \textbf{8.04} \\
    \midrule
    \multicolumn{8}{l}{\textit{Ablations}} \\
    \quad w/o agentic cleaning              & 7.84 & 8.12 & 7.32 & 7.69 & 7.55 & 7.83 & 7.71 \\
    \rowcolor{gray!10} \quad \footnotesize $\Delta$  & \footnotesize \textcolor[rgb]{0,0.5,0}{$-$0.26} & \footnotesize \textcolor[rgb]{0,0.5,0}{$-$0.22} & \footnotesize \textcolor[rgb]{0,0.5,0}{$-$0.29} & \footnotesize \textcolor[rgb]{0,0.5,0}{$-$0.38} & \footnotesize \textcolor[rgb]{0,0.5,0}{$-$0.28} & \footnotesize \textcolor[rgb]{0,0.5,0}{$-$0.29} & \footnotesize \textbf{\textcolor[rgb]{0,0.5,0}{$-$0.33}} \\
    \addlinespace[2pt]
    \quad w/o iterative composition          & 6.05 & 6.41 & 5.32 & 5.94 & 5.71 & 6.10 & 5.89 \\
    \rowcolor{gray!10} \quad \footnotesize $\Delta$ & \footnotesize \textcolor[rgb]{0,0.5,0}{$-$2.05} & \footnotesize \textcolor[rgb]{0,0.5,0}{$-$1.93} & \footnotesize \textcolor[rgb]{0,0.5,0}{$-$2.29} & \footnotesize \textcolor[rgb]{0,0.5,0}{$-$2.13} & \footnotesize \textcolor[rgb]{0,0.5,0}{$-$2.12} & \footnotesize \textcolor[rgb]{0,0.5,0}{$-$2.02} & \footnotesize \textbf{\textcolor[rgb]{0,0.5,0}{$-$2.15}} \\
    \bottomrule
    \end{tabular}
}
\end{table}

\noindent\textbf{Baseline comparison.}\ \Editor\ leads on every evaluation axis (Table~\ref{tab:editable_optimized}; Figure~\ref{fig:editor-qualitative}), achieving an overall score of $8.04$ against $6.91$ for AutoFigure-Edit and $3.69$ for Edit-Banana. The margin is widest on the structural axes (text and arrows) where precise coordinate reasoning and iterative correction matter most, as the examples in Figure~\ref{fig:editor-qualitative} confirm. Two complementary designs close this gap: the instruction-driven extraction phase resolves overlapping elements before composition and supplies the composer with clean per-element assets rather than noisy crops, while the iterative composition phase with its hybrid critic catches structural errors across refinement rounds, with programmatic checkers auditing arrow endpoints and element overlap at each iteration.

\noindent\textbf{Design effectiveness.}\ The ablation rows in Table~\ref{tab:editable_optimized} quantify each design's contribution. Removing iterative composition causes a sharp and uniform drop across all seven axes (overall $-2.15$), confirming that critic-driven revision is what turns a brittle one-shot SVG into a faithful reproduction. Removing agentic cleaning yields a smaller but consistent effect (overall $-0.33$), with the largest per-axis drop falling on icons, where clean extraction of overlapping visual assets is most critical. Together, both ablations confirms that the two designs are jointly necessary for the \Editor.
\section{Conclusion}
\label{sec:conclusion}

We have presented a harness-based approach to scientific figure authoring addressing two practical gaps left by existing systems: limited generalization across figure types and input conditions, and the inability to produce editable outputs. \Crafter\ and \Editor\ instantiate this harness for figure generation and raster-to-SVG conversion respectively, and \CrafterBench\ provides the first benchmark for cross-type, cross-condition evaluation. Experiments confirm that \Crafter\ outperforms all baselines on both \PaperBananaBench\ and \CrafterBench, that every mechanism contributes independently, and that \Editor\ leads prior raster-to-editable methods across all evaluation axes. Because the harness is executor-agnostic, stronger future generators can be incorporated without modification, and we expect the same pattern to extend to structured-output domains beyond scientific figures.

\bibliographystyle{plainnat}

\bibliography{references}

\appendix
\crefname{section}{Appendix}{Appendices}
\Crefname{section}{Appendix}{Appendices}
\renewcommand{\thefigure}{A\arabic{figure}}
\renewcommand{\thetable}{A\arabic{table}}
\setcounter{figure}{0}
\setcounter{table}{0}

\section{Appendix}
\noindent This appendix is organized as follows.

\begin{itemize}[leftmargin=1.4em,itemsep=2pt,topsep=2pt]
    \item In \Cref{app:experiment_setup}, we give the experimental setup for both harnesses: \Crafter\ on \PaperBananaBench\ and \CrafterBench\ (\Cref{app:setup-crafter}) and \Editor\ on raster-to-vector conversion (\Cref{app:setup-editor}).
    \item In \Cref{app:bench-details}, we detail \CrafterBench\ construction: source pools and crawl windows (\Cref{sec:bench-sources}), the quality gates (\Cref{app:bench-qc}), and the reference-conditioned task construction (\Cref{app:bench-edit-construction}).
    \item In \Cref{app:crafter-impl}, we provide \Crafter\ implementation details, including the scaling behavior of $K$ and $T$ (\Cref{app:scaling}) and the computational cost (\Cref{app:cost}).
    \item In \Cref{app:editable-ablations}, we present \Editor\ implementation details and ablations.
    \item In \Cref{app:editor-judge}, we describe the \Editor\ judge-ensemble protocol.
    \item In \Cref{app:eval-protocol}, we give the full \CrafterBench\ evaluation protocol.
    \item In \Cref{app:judge-prompts}, we reproduce the full judge prompts.
    \item In \Cref{app:human-eval}, we report a human evaluation validating the automatic judge.
    \item In \Cref{app:limitations}, we discuss limitations.
    \item In \Cref{app:cases}, we present additional qualitative case studies.
    \item In \Cref{app:failures}, we analyze representative failure cases of \Crafter.
\end{itemize}

\section{Experimental Setup}
\label{app:experiment_setup}

This appendix details the experimental setup summarized in Section~\ref{sec:experiments}.

\subsection{\Crafter\ on \PaperBananaBench\ and \CrafterBench}
\label{app:setup-crafter}

\noindent\textbf{Benchmarks and metric.}\
We evaluate on \PaperBananaBench~\citep{zhu2026paperbanana}, which covers text-to-image generation of $292$ academic methodology figures, and on our proposed \CrafterBench\ ($n{=}279$; Section~\ref{sec:bench}). \PaperBananaBench\ is scored under its official protocol with Gemini~3.1~Pro (\texttt{gemini-3.1-pro-preview}) as the VLM judge. \CrafterBench\ is scored under the per-image referenced protocol of Section~\ref{sec:bench-judge} and Appendix~\ref{app:eval-protocol} with Gemini~3.5~Flash. Both report the lenient win-rate, the average of the per-sample $\{100, 50, 0\}$ verdict mapping.

\noindent\textbf{Baselines.}\
We compare against three groups of methods: two open-source generators (GLM-Image~\citep{glm2025image}, Qwen-Image~\citep{qwen2025image}), three closed-source generators (GPT-Image-2~\citep{openai2025gptimage2}, Nano~Banana~2~\citep{google2025nanobanana2}, Nano~Banana~Pro~\citep{google2025nanobananapro}), and two agentic frameworks (PaperBanana~\citep{zhu2026paperbanana} and AutoFigure~\citep{zhu2026autofigure}). Vanilla generators receive the same caption and source paper-text as the harness pipelines and are queried once with the model's default decoding parameters, with no additional retries.

\noindent\textbf{Controlled comparison.}\
To isolate the effect of orchestration design, all agentic methods share the same image-generation backbone (Nano~Banana~2) and the same vision-language model (\texttt{gemini-3.1-pro-preview}) for all vision-dependent agents. We additionally report \Crafter\ with Nano~Banana~Pro to verify executor pluggability (Section~\ref{sec:method-harness}). Full model assignments and per-agent configurations are in Appendix~\ref{app:crafter-impl}.

\noindent\textbf{Inference and judging.}\
Each benchmark is run end-to-end once per method.
\PaperBananaBench\ uses the official per-dimension judging
configuration of \citet{zhu2026paperbanana}. On \CrafterBench, each
image is scored once per applicable aspect using the prompts of
Appendix~\ref{app:judge-prompts}, with the judge queried at
\texttt{temperature\,=\,0} under a fixed seed and up to $8$ retries
with exponential backoff. A missing or corrupt generation counts as
an automatic \textit{Human} verdict and is never dropped.
Random seeds for the harness loop are fixed across baselines so
that the agentic frameworks see the same per-sample plan
candidates.

\subsection{\Editor\ on Raster-to-Vector Conversion}
\label{app:setup-editor}

\noindent\textbf{Subset.}\
\Editor\ is evaluated on a random subset of $80$ rasters drawn from \Crafter's outputs across both \PaperBananaBench\ and \CrafterBench, balanced across academic, poster, and infographic figure types so that each type contributes a comparable share. The subset is held out of the main \Crafter\ comparison and is used only for the editable-output evaluation.

\noindent\textbf{Baselines.}\
We compare against two raster-to-vector systems designed for making automated figure outputs editable: Edit-Banana~\citep{editbanana2026}, which converts the raster through a SAM-3 segmentation pipeline and writes DrawIO cells, and AutoFigure-Edit~\citep{lin2026autofigureedit}, which detects icons with SAM-3 and emits a full SVG in a single LLM call. Both baselines receive the same input raster as \Editor\ and are run with their public default settings.

\noindent\textbf{Judge ensemble.}\
A three-VLM ensemble scores each output on seven axes (\textit{position}, \textit{color}, \textit{text}, \textit{icon}, \textit{arrow}, \textit{style}, plus a holistic \textit{overall}) on a $0$--$10$ scale. The three judges are Gemini~3.1~Flash-Lite, GPT-5, and Doubao-Seed-2.0-Pro, queried independently with the same prompt and aggregated by mean. A single-judge retry rule re-queries any model returning an overall score below $3.0$ once; the retry replaces the original when its overall is higher. Full prompt, aggregation rule, and per-model versions are in Appendix~\ref{app:editor-judge}.

\noindent\textbf{Ablations.}\
The two \Editor\ ablations (no agentic cleaning; no iterative composition) are run on the same $80$-sample subset with the same three-VLM ensemble; per-stage parameter changes and ablation prompts are in Appendix~\ref{app:editable-ablations}.

\section{\CrafterBench: Dataset Details}
\label{app:bench-details}

\subsection{Source Pool Composition and Crawl Windows}
\label{sec:bench-sources}

The $279$ samples in \CrafterBench\ are drawn from five source pools (Table~\ref{tab:bench-sources}). The academic pool is split across two arXiv crawls: a broad-domain crawl targeted at general method-figure coverage, and a method/architecture crawl targeted at the figure types that underpin the key-element-composition and sketch-conditioned-generation constructions (figures whose caption contains \textit{overview}, \textit{pipeline}, \textit{architecture}, \textit{method}, \textit{approach}, or \textit{framework}).

\begin{table}[h]
\centering
\caption{Source pools feeding \CrafterBench. Each surviving sample passes the seven-stage quality pipeline of Appendix~\ref{app:bench-qc}; NeurIPS spotlight + oral posters were also crawled but none survived the gates.}
\label{tab:bench-sources}
\resizebox{\linewidth}{!}{
\begin{tabular}{l r l}
\toprule
\textbf{Pool} & \textbf{n} & \textbf{Provenance} \\
\midrule
arXiv (broad-domain crawl)                  &  84 & cs.LG / cs.CV / cs.CL / cs.RO / cs.AI / cs.NE / cs.HC / cs.IR / \\
                                            &     & cs.GR / cs.SD / cs.DB / cs.MM / cs.CR / eess.IV / eess.AS / \\
                                            &     & stat.ML / q-bio (BM,NC,QM) / physics (bio-ph, med-ph) \\
arXiv (method/architecture crawl)           &  56 & method-figure crawl restricted to overview / pipeline / \\
                                            &     & architecture / method / approach / framework captions \\
\midrule
CVPR posters (highlight + award)            &  55 & CVPR highlight and award tier \\
\midrule
ICLR posters (oral + spotlight)             &  54 & ICLR oral and spotlight via OpenReview \\
\midrule
Lil'Log (Lilian Weng's blog)                &  30 & deep-dive ML topics \\
\midrule
\textbf{Total}                              & \textbf{279} & \\
\bottomrule
\end{tabular}
}
\end{table}

\subsection{Quality Gates}
\label{app:bench-qc}

Every candidate passes a seven-stage pipeline before entering \CrafterBench:
\begin{enumerate}[leftmargin=1.5em, itemsep=0pt, topsep=0pt, parsep=0pt]
    \item \textbf{Caption keyword filter} (G1): requires figure-type language and method-related keywords (e.g., \textit{overview}, \textit{architecture}, \textit{pipeline}) in the caption.
    \item \textbf{Strict content classifier} (G2): a vision-language classifier assigns each figure to one of 15 fine-grained types; only \textit{diagram}, \textit{illustration showing method}, \textit{architecture}, and \textit{teaser} are accepted. Photographs, statistical charts, screenshots, equation-only renders, and tables are rejected.
    \item \textbf{Complexity rescore} (G3): a vision-language rubric verifies that the figure is worth recreating as a drawing, exhibits sufficient design richness (score $\geq 4/5$), contains at least $8$ distinct named components, and would take an estimated $10$+ minutes to recreate manually.
    \item \textbf{Alignment verification} (G4): a vision-language check verifies that at least $70\%$ of authored visual claims match the figure content, at least $60\%$ of proposed edit targets are feasible, and the caption alignment score reaches $\geq 3/5$.
    \item \textbf{First-pass quality assurance} (G5): a vision-language reviewer flags cropping artifacts, watermarks, low resolution, caption mismatch, and edit-target referent issues.
    \item \textbf{Evidence-required quality assurance} (G6): a stricter second pass in which every flag must cite direct pixel-region evidence with confidence $\geq 4/5$; unsupported flags are auto-discarded.
    \item \textbf{Manual review} (G7): human inspection of every flagged sample plus all reference-conditioned task inputs. The interface and acceptance rule are described in Section~\ref{app:bench-edit-construction}.
\end{enumerate}

\subsection{Reference-Conditioned Task Construction}
\label{app:bench-edit-construction}

For each of the three reference-conditioned tasks, conditioning inputs are constructed from the source figures through a semi-automatic pipeline followed by manual quality assurance. Every input is inspected and either hand-edited or hand-confirmed before it enters the benchmark, and the counts below reflect samples that survived the full quality pipeline including unanimous human agreement.

\noindent\textbf{Mask-completion} ($n{=}30$).\
A vision-language model proposes a semantic region whose removal still leaves the figure interpretable, and the region is masked to pure white so that the unmasked pixels stay identical to the ground truth. The region's identity is recorded as a short natural-language label whose wording is hand-polished for every sample. Mask area ranges from roughly $20\%$ to over $90\%$ of the figure, averaging about $40\%$. The masks span three layouts (single rectangle $12$, multiple rectangles $10$, single unlabeled region $8$), with $27$ inputs hand-edited by re-cropping the bounding box or repainting the region and $3$ hand-confirmed without modification.

\noindent\textbf{Key-element composition} ($n{=}30$).\
An image-editing model extracts only the icon-level placeholders that define the figure's spatial logic into a minimal, representative element skeleton, with text labels and connecting arrows stripped. Small random displacement is applied to each extracted element to discourage trivial gap-filling. A human annotator reviews each result, deleting incorrectly extracted elements and equalizing difficulty ($29$ hand-edited, $1$ hand-confirmed).

\noindent\textbf{Sketch-conditioned generation} ($n{=}40$).\
A layout reference is prepared in one of three sketch families: hand-drawn or pen sketches ($n{=}15$), AI-drafted rough sketches conditioned on the caption and surrounding context, deliberately not on source pixels, to keep the sketch genuinely rough rather than a polished retrace ($n{=}14$), and SVG wireframes generated from the source figure's vector code and rasterized ($n{=}11$). Every sketch input passes human quality assurance.

\begin{figure}[!htbp]
\centering
\includegraphics[width=0.5\linewidth]{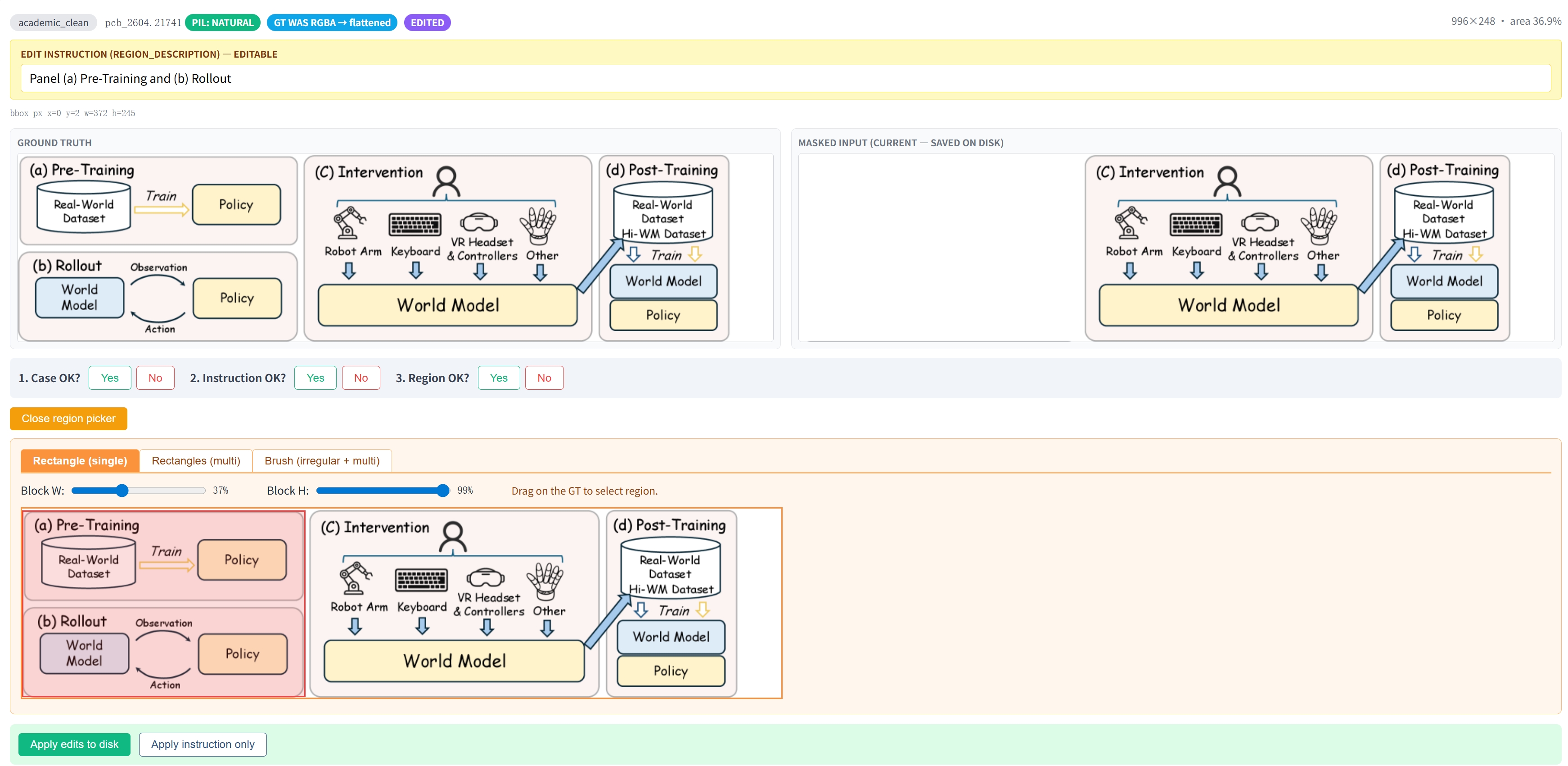}\\
{\small (a) Mask-completion}\\[0.4em]
\includegraphics[width=0.5\linewidth]{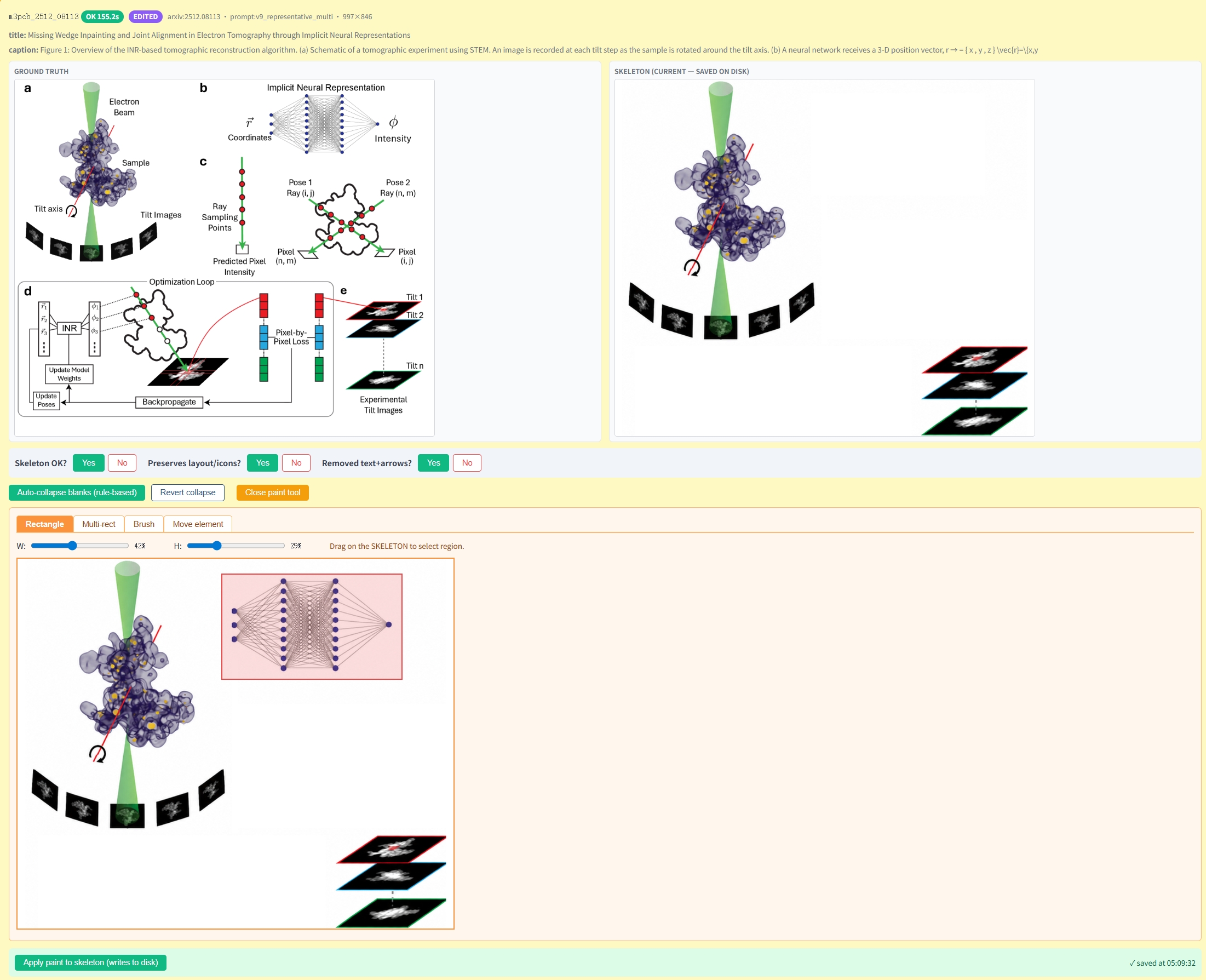}\\
{\small (b) Key-element composition}\\[0.4em]
\includegraphics[width=0.5\linewidth]{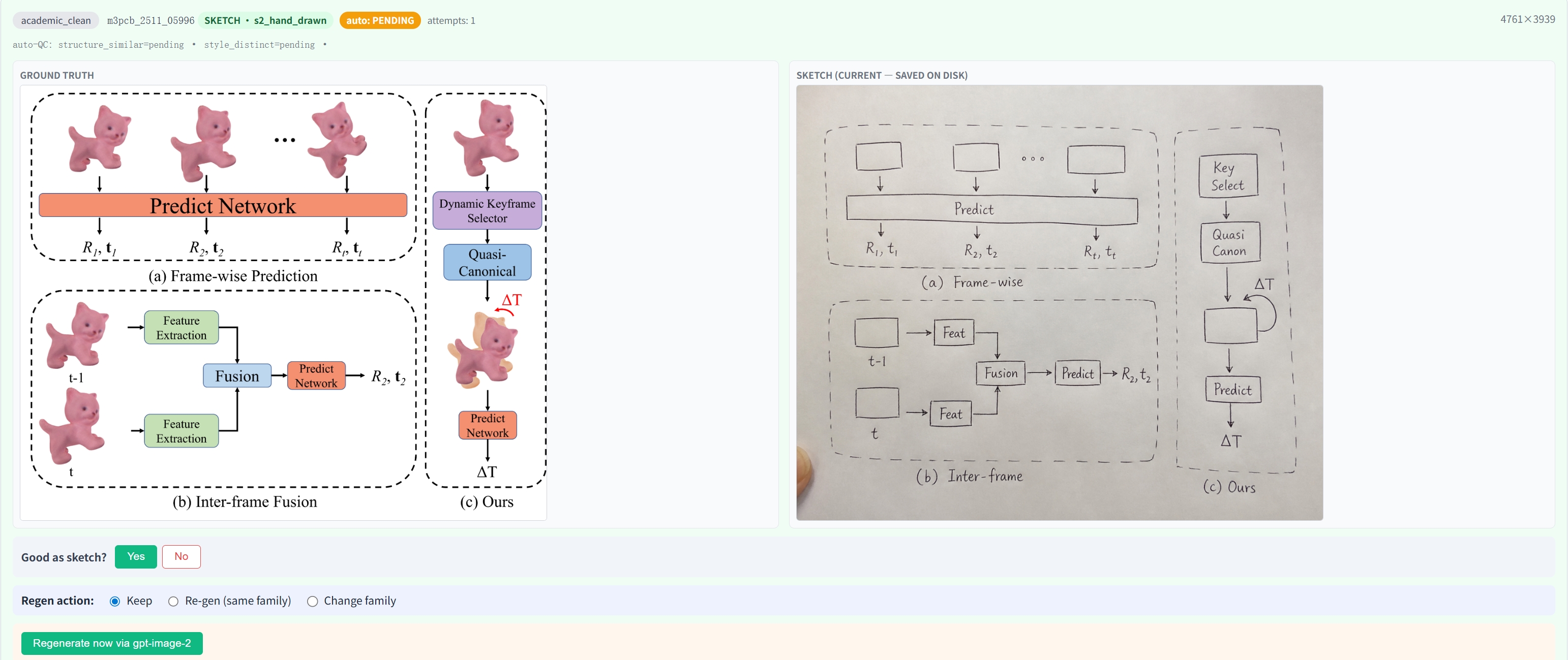}\\
{\small (c) Sketch-conditioned generation}
\caption{Manual annotation interfaces for the three reference-conditioned task constructions. Each interface combines the auto-generated conditioning input with task-specific drawing tools (drag-to-mask overlay, rectangle/brush erasers, family-switch regeneration); three graduate-level annotators inspect every sample and a unanimous-agreement rule decides acceptance.}
\label{fig:annotation-tools}
\end{figure}

\noindent\textbf{Manual quality assurance.}\
Every reference-conditioned task input is reviewed through a dedicated browser annotation interface, one per task family, with three graduate-level annotators independently inspecting each sample. A sample is accepted into \CrafterBench\ only when all three annotators agree it is acceptable; any disagreement triggers a revision (hand-recrop, hand-paint, regenerate, or family swap) followed by re-review until unanimous agreement is reached. The three interfaces and their per-task operations are summarized in Figure~\ref{fig:annotation-tools}: mask-completion uses a drag-to-mask overlay with an editable region label; key-element composition uses rectangle, multi-rectangle, and brush erasers to remove leftover text and arrows after automatic extraction; sketch-conditioned generation uses the same erasers plus a one-click family-switch that redraws the layout reference under a different sketch family when the current one is rejected.

\section{\Crafter: Implementation Details}
\label{app:crafter-impl}

\noindent\textbf{Model configuration.}\
For fair comparison, all agentic methods share the same image-generation backbone and vision-language model. The default backend $\mathcal{E}$ is Nano~Banana~2 (\texttt{gemini-3.1-flash-image-preview}); we additionally report \Crafter\ with Nano~Banana~Pro (\texttt{gemini-3.0-pro-image-preview}). Vision-dependent agents (critic $\mathcal{V}$, convergence judge) use \texttt{gemini-3.1-pro-preview} across \Crafter, PaperBanana, and AutoFigure alike. For the language-only agents (intent reasoner, plan generator $\mathcal{D}$, specification refiner $\mathcal{R}$), \Crafter\ adopts \texttt{claude-opus-4-6}~\citep{claudeopus46}. For the open-source baselines, GLM-Image is evaluated directly; Qwen-Image uses Qwen-Image-2512 for text-to-image and Qwen-Image-Edit-2511 for reference-conditioned tasks.

\noindent\textbf{Pipeline orchestration.}\
The harness instantiates the five agents coordinated around the evolving figure specification $\mathcal{S}$ (Section~\ref{sec:method-harness}). $\mathcal{E}$ supplies raster outputs but is not itself an agent. All agents read and write $\mathcal{S}$; free-text addenda from one agent never reach another agent's prompt directly. $\mathcal{D}$ emits $K \in \{1,2,3\}$ visual-style keys from an $11$-key vocabulary (banner, multi-column grid, numbered-step sequence, etc.); the convergence judge bounds the loop at $T{=}3$ rounds; the early-exit gate fires when content accuracy and role conformity both exceed $6.5/10$ and no pixel artifact is detected.

\noindent\textbf{Image-generation routing.}\
For text-to-image samples, $\mathcal{E}$ is called once per plan \citep{si2023spokenwoz, si2026goalplanjustwish} with a single text prompt. For reference-conditioned samples, $\mathcal{E}$ is called via the multimodal interface with the reference image as the first item and a task-specific instruction as the text. Adding a new reference-conditioned task requires one additional branch in the instruction builder plus one entry in the reference-image role-hint table; no other pipeline code changes.

\noindent\textbf{Typed corrective edits.}\
$\mathcal{R}$ emits typed edits to $\mathcal{S}$ from a fixed vocabulary of structured operations (adding layout constraints, banning artifact categories, resizing or demoting named elements). Each edit composes through $\mathcal{S}$ rather than as a free-text prompt addendum, so the next round's prompt builder sees a coherent typed delta. $\mathcal{V}$ emits a directive diagnostic $d_t$ containing per-dimension scores along six quality axes (content accuracy, layout coherence, text legibility, role conformity, aesthetic quality, artifact severity), an issues list, a suggestions list, and a revised figure description; $\mathcal{R}$ reads only the issues and suggestions to decide which edits to apply.

\noindent\textbf{Convergence judge.}\
The convergence judge applies hard rules (iteration cap, content-score threshold, overall-score threshold, pixel-artifact detection) when applicable, and falls back to a vision-language acceptance call otherwise. After the loop terminates, the judge selects the highest-scoring artifact $a^*$ across all rounds, with a post-correction pass (OCR-based typo repair on rendered text, plus a quality-guard revert when post-correction introduces an artifact).

\subsection{Scaling Behavior of $K$ and $T$}
\label{app:scaling}

We vary the number of candidate plans $K$ and refinement rounds $T$ independently on \PaperBananaBench\ to understand each dimension's contribution. Results are shown in Table~\ref{tab:scaling-kt}.

\begin{table}[h]
\centering
\small
\caption{Effect of plan candidates $K$ and refinement rounds $T$ on \PaperBananaBench. Each group varies one factor while holding the other fixed. $\Delta$: gap vs.\ full configuration (last row).}
\label{tab:scaling-kt}
\begin{tabular}{@{}ll ccccc r@{}}
\toprule
& & \textbf{Faith.} & \textbf{Conc.} & \textbf{Read.} & \textbf{Aesth.} & \textbf{Overall} & \textbf{$\Delta$} \\
\midrule
\multicolumn{8}{l}{\emph{Varying $K$ (fixed $T{=}3$)}} \\
$K{=}1$        && 28.42 & 51.20 & 38.30 & 60.10 & 41.78 & $-$8.56 \\
$K{=}3$        && 34.25 & 56.16 & 48.45 & 66.44 & 48.97 & $-$1.37 \\
\midrule
\multicolumn{8}{l}{\emph{Varying $T$ (fixed $K{=}\text{adaptive}$)}} \\
$T{=}1$        && 30.07 & 51.97 & 41.55 & 61.80 & 44.86 & $-$5.48 \\
\midrule
\rowcolor{blue!5}
\multicolumn{2}{l}{\textbf{Full} ($K{=}\text{adaptive},\; T{=}3$)} & 38.18 & 53.42 & 47.77 & 64.21 & 50.34 & \\
\bottomrule
\end{tabular}
\end{table}

Increasing $K$ from $1$ to $3$ yields the largest single gain ($+7.19$~point), confirming that plan-level diversity is critical for escaping structurally unsuitable framings. Moving from fixed $K{=}3$ to adaptive $K$ adds a further $+1.37$~point; the adaptive strategy allocates more candidates to complex, multi-constraint inputs and fewer to simple ones, producing a substantial faithfulness improvement ($+3.93$~point over $K{=}3$) on the samples where content correctness is hardest. Increasing $T$ from $1$ to $3$ yields $+5.48$~point, confirming that iterative refinement provides consistent returns. Together, plan diversity and iterative refinement contribute complementary gains: $K$ determines whether the pipeline starts from a viable framing, while $T$ determines whether localized errors in that framing get corrected.

\subsection{Computational Cost}
\label{app:cost}

Table~\ref{tab:cost} reports the per-figure inference cost for \Crafter, PaperBanana, and \Editor.

\begin{table}[h]
\centering
\small
\caption{Per-figure generation cost.}
\label{tab:cost}
\begin{tabular}{@{}lr@{}}
\toprule
\textbf{System} & \textbf{Cost} \\
\midrule
AutoFigure (w/ Nano Banana 2) & \$0.06 \\
PaperBanana (w/ Nano Banana 2) & \$0.11 \\
\Crafter\ (w/ Nano Banana 2) & \$0.25 \\
\Crafter\ (w/ Nano Banana Pro) & \$0.32 \\
\midrule
\Editor\ (per conversion) & \$0.85 \\
\bottomrule
\end{tabular}
\end{table}

\Crafter\ costs approximately $2$--$3\times$ more per figure than 
PaperBanana, trading inference budget for multi-variant plan 
exploration and iterative refinement. This overhead is modest in 
context: a publication-quality figure typically requires hours of 
manual effort, and the total cost of generating all $279$
\CrafterBench\ samples is under \$90. \Editor\ adds \$0.85 per 
raster-to-SVG conversion, with the majority of cost attributable to 
LLM output tokens during iterative SVG refinement.

\section{\Editor: Implementation Details and Ablations}
\label{app:editable-ablations}

\noindent\textbf{Extraction phase.}\ 
A vision-language designer agent $\mathcal{D}$ inspects the input
raster $a^*$ and authors a per-figure keep/delete plan; an
instructable image-editing executor $\mathcal{E}$ carries out the
plan at the pixel level.
A verify-then-refine wrapper runs at most $T{=}3$ iterations: a
verifier $\mathcal{V}$ (a lightweight VLM, decoupled from the
editor) inspects each cleaned candidate and either accepts it or
returns a directive diagnostic (e.g., ``the bottom-row icons were
over-deleted; restore them; remove the page number instead'') that
triggers another round.
Iteration-count distribution on dense posters: $47\%$ converge at
round~1, $46\%$ at round~2, $7\%$ at round~3.

\noindent\textbf{Composition phase.}\ 
After the processing phase (captioning, referring grounding, and
per-element vector/raster classification), a language model
generates two candidate SVG skeletons at decoding temperatures
$0.20$ and $0.45$; the convergence judge picks the better
candidate.
Extracted assets are spliced into the placeholders, and $T{=}4$
refinement rounds run with the hybrid critic reporting per-axis
scores (text presence, arrow endpoints, layout consistency, color
drift).
A best-so-far reversion guards against non-monotonic regressions:
without it, approximately $30\%$ of refinement iterations score
lower than the immediately preceding iteration.

\noindent\textbf{Provider abstraction.}\ 
Four external services (LLM, image editor, segmentation, background
removal) are wrapped behind interface adapters.
Swapping a backend (e.g., the segmentation model for ablation) is a
single configuration change.

\noindent\textbf{Ablation setup.}\ 
Two ablations target the two harness-instantiation phases of
\Editor\ (Table~\ref{tab:editable_optimized}), evaluated on the
same $80$-sample subset with the same three-VLM judge ensemble
(Appendix~\ref{app:editor-judge}).
\emph{w/o agentic cleaning}: the extraction phase (Stage~1) is
skipped; captioning runs on the original raster and per-element
extraction falls back to a segmentation-plus-background-removal
alternative.
\emph{w/o iterative composition}: the composition-phase refinement
loop is disabled ($T{=}0$); the skeleton with injected assets is
returned directly without critic-driven revision.

\noindent\textbf{Per-category extraction-phase analysis.}\ 
\Editor\ wins or ties the ``w/o agentic cleaning'' ablation in $11$
of $12$ source categories under the headline judge ensemble.
The single exception is a $3$-sample text-to-image infographic
subset where the absolute magnitudes fall within the per-sample
noise band; the extraction-phase signal is statistically
inconclusive on this subset.
\section{\Editor: Judge Ensemble Protocol}
\label{app:editor-judge}

\noindent\textbf{Models and prompt.}\
The judge ensemble comprises three independent VLMs: Gemini~3.1~Flash-Lite, GPT-5.4, and Doubao-Seed-2.0-Pro. Each judge receives the original input raster and the rendered SVG as two image attachments and scores seven axes (\emph{position}, \emph{color}, \emph{text}, \emph{icon}, \emph{arrow}, \emph{style}, \emph{overall}) on a $0$--$10$ scale, returning a JSON response with per-axis scores and a structured issues list. All calls use temperature $0.15$ and a $4{,}000$-token output cap.

\noindent\textbf{Aggregation.}\
The headline score per sample is the mean of the three judges' \emph{overall} scores. Any judge returning an \emph{overall} score below $3.0$ is re-queried once; if the retry scores higher, it replaces the original. This retry rule mitigates known VLM-judge volatility on visually unfamiliar inputs.

\noindent\textbf{Relationship to the generation rubrics.}\
The seven-axis editable-output rubric is distinct from the generation rubrics used on \PaperBananaBench\ and \CrafterBench\ (Section~\ref{sec:bench-judge}). Those rubrics score generation quality against a human-drawn target, whereas the seven-axis rubric scores reproduction fidelity of a raster-to-SVG conversion against the input raster. The two are reported in separate tables (Table~\ref{tab:main-results} vs.\ Table~\ref{tab:editable_optimized}) and should not be compared directly.

\section{Evaluation Protocol Details}
\label{app:eval-protocol}

We score \CrafterBench\ with a per-image referenced protocol. For each sample, a Gemini~3.5~Flash judge (temperature $0$, fixed seed) scores the generated figure and the human-drawn ground truth independently, one image at a time, against the same inputs the generator received (caption, paper context, and, for the reference-conditioned tasks, the input image). Scoring each image on its own rather than as an A/B pair avoids position bias. Each image is rated from $0$ to $10$ on a small aspect set chosen by task and content type: content faithfulness and readability on every sample, plus a style format aspect for text-to-image (academic, poster, or infographic) or an input-fidelity aspect for the three reference-conditioned tasks. The exact aspect prompts are reproduced in Appendix~\ref{app:judge-prompts}.

A weighted mean of the aspect scores gives each image a single total, weighting content faithfulness and input fidelity most heavily ($3.0$) and readability and format least ($1.0$ to $1.5$). The generated figure is judged \textit{Model} when its total exceeds the ground truth's by more than a tie band of $0.30$, \textit{Human} when it trails by more than $0.30$, and \textit{Tie} otherwise. A missing generation counts as \textit{Human}. The bench-level score is the lenient win-rate, the mean over samples of the $\{100, 50, 0\}$ mapping of these verdicts, and on academic text-to-image inputs it reduces to a PaperBanana-style referenced judge. \PaperBananaBench\ is scored separately under its official protocol with Gemini~3.1~Pro.

\section{Judge Prompts}
\label{app:judge-prompts}

The \CrafterBench\ judge runs as four jobs, one per task type (text-to-image, key-element composition, sketch-conditioned generation, mask-completion). All four share the system scaffold and scoring anchors of the first box below. The per-sample aspect set ($\{$\texttt{aspects}$\}$) is drawn from the second box, and the user message is filled from the third. Text blanks $\{\{\dots\}\}$ are filled from sample metadata, and image blanks $[[\dots]]$ are attached as JPEG.

\begin{figure}[t]
\centering
\begin{tcolorbox}[title={System prompt and scoring anchors (shared by all four jobs, with the edit jobs additionally foregrounding input\_fidelity)}, width=\textwidth, colback=gray!5, colframe=black, fonttitle=\bfseries]
\small
You are a STRICT, skeptical reviewer of scientific figures. You will see the \\
original request given to an illustrator, optionally an INPUT IMAGE they were \\
told to work from, and ONE candidate figure. One of the candidates you review \\
is a real human-made figure from a published paper/poster/blog; the other is \\
AI-generated. AI figures often look glossy yet hide subtle flaws: garbled or \\
fake text, plausible-but-wrong components, generic placeholders, or (for edit \\
tasks) silently ignoring/altering the input image. Hunt for these. \\[5pt]
Score each aspect 0-10 with CONSERVATIVE anchors: \\
\hspace*{1.0em}9-10 = flawless on this aspect after careful search (you found NO weakness) \\
\hspace*{1.0em}7-8  = good, minor issues \\
\hspace*{1.0em}5-6  = acceptable but clearly flawed \\
\hspace*{1.0em}3-4  = major problems \\
\hspace*{1.0em}0-2  = fails / unrelated / garbled \\
Do NOT default to 9-10. Most figures have at least minor flaws; if you are \\
about to give 9-10, re-examine and try to find a weakness first. Reward \\
correctness and communication, not surface polish. \\[5pt]
Aspects to score for THIS sample: \\
\{aspects\} \\[5pt]
First list concrete weaknesses you actually observe (be specific; "none" only \\
if truly flawless), then score. Return STRICT JSON only: \\
\{"weaknesses": ["<specific flaw>", ...], "scores": \{<aspect keys>\}\}
\end{tcolorbox}
\caption{\CrafterBench\ judge: shared system prompt and scoring anchors. The edit jobs additionally foreground input fidelity.}
\label{fig:prompt-judge-system}
\end{figure}

\begin{figure}[t]
\centering
\begin{tcolorbox}[title={Aspect definitions filling the $\{$aspects$\}$ slot. Text-to-image uses content\_faithfulness, readability, and one format aspect, while the edit tasks use content\_faithfulness, readability, and input\_fidelity.}, width=\textwidth, colback=gray!5, colframe=black, fonttitle=\bfseries]
\small
content\_faithfulness \\
\hspace*{1.0em}Does it convey the content in the caption/brief ACCURATELY and COMPLETELY? \\
\hspace*{1.0em}Check the specific components, relationships and flow named in the caption \\
\hspace*{1.0em}are present and correct. CRITICAL: scrutinize the text in the figure: are \\
\hspace*{1.0em}labels REAL, specific and correct, or are some garbled / nonsensical / vague \\
\hspace*{1.0em}placeholders / AI-gibberish? Are any components fabricated or missing? \\
\hspace*{1.0em}Plausible-looking but wrong = low. 9-10 only if every caption element is \\
\hspace*{1.0em}present, correct, and all text is genuine and legible. \\[5pt]
readability \\
\hspace*{1.0em}Is the information easy to extract: legible text, clear visual flow, sensible \\
\hspace*{1.0em}grouping, no clutter/overlap/spaghetti arrows? 9-10 only if a reader \\
\hspace*{1.0em}navigates it effortlessly. \\[5pt]
format\_academic | format\_poster | format\_infographic   [t2i: one, by content type] \\
\hspace*{1.0em}academic    = clean single-figure diagram, restrained styling, crisp \\
\hspace*{8.0em}components/arrows; no poster banner, no infographic decoration. \\
\hspace*{1.0em}poster      = real conference poster: title banner, author/affiliation line, \\
\hspace*{8.0em}>=3 sections with headers, rich content; penalize a thin skeleton. \\
\hspace*{1.0em}infographic = casual explainer (Distill/Quanta/blog style): illustrative \\
\hspace*{8.0em}icons or a visual metaphor, narrative callouts, friendly \\
\hspace*{8.0em}non-academic typography; penalize a dry academic block diagram. \\[5pt]
input\_fidelity   [edit tasks only; judged OUTPUT vs INPUT IMAGE only] \\
\hspace*{1.0em}key-element : are the SPECIFIC provided elements actually REUSED (recognizably \\
\hspace*{8.0em}the same icons/photos/charts), not replaced by similar-but- \\
\hspace*{8.0em}different ones? Ignoring or substituting them MUST score 0-4; \\
\hspace*{8.0em}8-10 only if most/all provided elements are clearly reused. \\
\hspace*{1.0em}sketch      : scores (a) LAYOUT-FOLLOWING (keep the sketch's panels, labels, \\
\hspace*{8.0em}arrow connections, reading order) and (b) POLISH (a genuinely \\
\hspace*{8.0em}refined publication figure, NOT a near-copy of the rough \\
\hspace*{8.0em}sketch). A near-copy MUST score 0-4 even if the layout matches; \\
\hspace*{8.0em}8-10 only if faithful AND clearly polished. \\
\hspace*{1.0em}mask-compl. : scores (a) SUBSTANTIVE FILL (the blank region is filled with \\
\hspace*{8.0em}appropriate content; leaving it empty or reproducing the input \\
\hspace*{8.0em}MUST score 0-3) and (b) PRESERVATION (the rest is unchanged); \\
\hspace*{8.0em}8-10 only if the fill is correct AND the rest is preserved.
\end{tcolorbox}
\caption{\CrafterBench\ judge: aspect definitions that fill the $\{$aspects$\}$ slot. Text-to-image uses content faithfulness, readability, and one format aspect; the edit tasks use content faithfulness, readability, and input fidelity.}
\label{fig:prompt-judge-aspects}
\end{figure}

\begin{figure}[t]
\centering
\begin{tcolorbox}[title={User message (text blanks filled from sample metadata; image blanks attached as JPEG)}, width=\textwidth, colback=gray!5, colframe=black, fonttitle=\bfseries]
\small
\#\#\# Text-to-image \\
ORIGINAL REQUEST \\
Task: t2i  |  Requested figure style: \{\{STYLE\}\} \\
Target figure caption: \\
\{\{CAPTION\}\} \\
SOURCE PAPER CONTEXT (judge faithfulness against this): \\
\{\{PAPER\_CONTEXT\}\} \\
Illustrator's brief: \\
\{\{INSTRUCTION\}\} \\
Candidate figure to score: \\
{}[[CANDIDATE\_IMAGE]] \\[5pt]
\#\#\# Edit tasks (key-element / sketch / mask-completion) \\
Target figure caption: \{\{CAPTION\}\} \\
SOURCE PAPER CONTEXT (content correctness ONLY; does NOT affect input\_fidelity): \\
\{\{PAPER\_CONTEXT\}\} \\
INPUT IMAGE given to the author (a brief to follow/fill, NOT to reproduce): \\
{}[[INPUT\_IMAGE]] \\
{}[mask-completion only] The blanked region should contain: \{\{REGION\_DESCRIPTION\}\} \\
Result to score (compare its structure/elements to the INPUT above): \\
{}[[CANDIDATE\_IMAGE]]
\end{tcolorbox}
\caption{\CrafterBench\ judge: user message template. Text blanks $\{\{\dots\}\}$ are filled from sample metadata and image blanks $[[\dots]]$ are attached as JPEG.}
\label{fig:prompt-judge-user}
\end{figure}

\section{Human Evaluation}
\label{app:human-eval}

\begin{figure}[t]
\centering
\includegraphics[width=\linewidth]{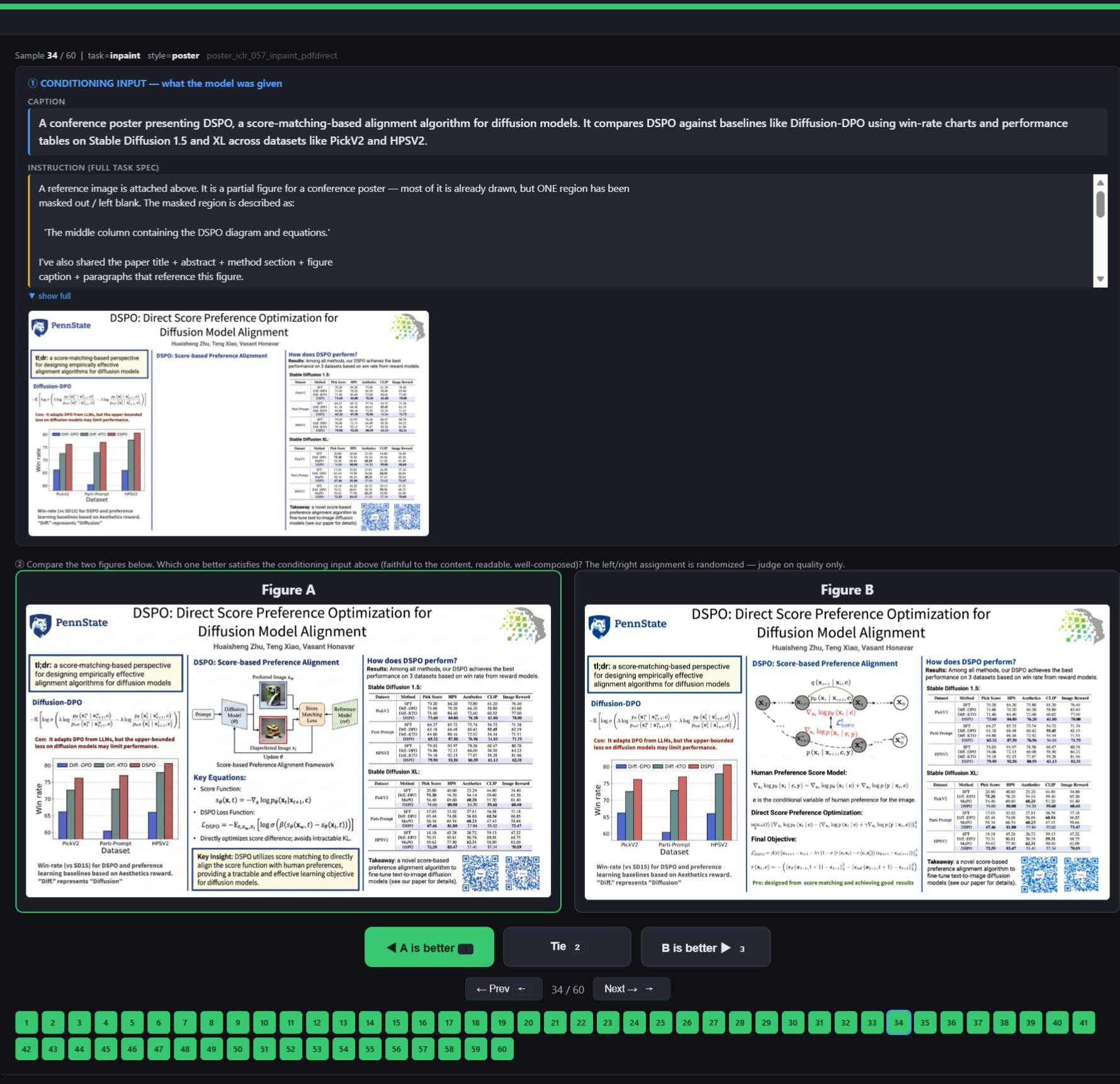}
\caption{The blind pairwise human-evaluation interface. The conditioning input (caption, full task instruction, and the input image for edit tasks) is shown at the top, and the annotator then chooses which of the two randomized figures better satisfies it.}
\label{fig:human-eval}
\end{figure}

Because the \CrafterBench\ score is produced by a VLM judge, we verify that it reflects \emph{human} preference with a blind pairwise study. Three graduate-level student annotators, compensated at a local rate of \$25 per hour, each rate a random sample of $60$ cases spanning all four tasks and three figure types. For each case, an annotator compares the model output against the original human-drawn figure through the custom web interface of Figure~\ref{fig:human-eval}, where the two are shown as ``Figure~A'' and ``Figure~B'' in a randomized, hidden order under the same conditioning input, and selects \emph{A is better}, \emph{Tie}, or \emph{B is better}. The instruction given to annotators is reproduced below.

\begin{figure}[t]
\centering
\begin{tcolorbox}[title={Instruction shown to annotators in the blind pairwise study}, width=\textwidth, colback=gray!5, colframe=black, fonttitle=\bfseries]
\small
You are shown the conditioning input the model was given (a caption, and for \\
edit tasks an input image such as a partial figure, an element collage, or a \\
rough sketch) and two rendered figures, A and B. One is the model output and \\
one is the original human-drawn figure, in randomized and hidden order. Choose \\
which figure better satisfies the request, namely faithful to the described \\
content, readable, well-composed, and (for edit tasks) consistent with the \\
provided input image, or Tie. Judge on quality only.
\end{tcolorbox}
\caption{Instruction shown to annotators in the blind pairwise human-evaluation study.}
\label{fig:prompt-human-eval}
\end{figure}

Mapping each comparison to a $\{\textsc{Model},\textsc{Tie},\textsc{Human}\}$ verdict and comparing it against our automatic judge, the metric agrees with the majority human verdict on $72\%$ of cases at Cohen's $\kappa = 0.58$ (Table~\ref{tab:human-eval}), confirming that the automatic score is a reliable proxy for human judgment across the four tasks and three figure types.

\begin{table}[H]
\centering
\small
\setlength{\tabcolsep}{12pt}
\renewcommand{\arraystretch}{1.1}
\caption{Human study ($N{=}60$ blind pairwise comparisons on a random sample, three annotators, spanning all four tasks and three figure types). The \CrafterBench\ judge tracks the majority human verdict.}
\label{tab:human-eval}
\resizebox{0.6\linewidth}{!}{%
\begin{tabular}{lcc}
\toprule
Automatic judge & Agreement (\%) & Cohen's $\kappa$ \\
\midrule
\CrafterBench\ judge (ours) & $72$ & $0.58$ \\
\bottomrule
\end{tabular}}
\end{table}

\section{Limitations}
\label{app:limitations}

Our headline numbers rely on closed-source models for both image-generation backbones (Gemini~3.1~Flash~Image, with comparisons to Gemini~3.0~Pro~Image and \texttt{openai/gpt-image-2}) and evaluation judges (Gemini~3.1~Pro for \PaperBananaBench, Gemini~3.5~Flash for \CrafterBench), making the harness contribution conditional on proprietary-model access and judge biases; \Crafter\ and \Editor\ further rely on a strong language model treated as a black box.
Per-sample latency and cost are non-trivial: a single \Crafter\ run executes up to four parallel image generations followed by up to three refinement rounds, and \Editor\ adds roughly four agentic VLM rounds plus an SVG composition step, so deploying the harness at scale requires the corresponding API budget and wall-clock time. 
Finally, \CrafterBench's $279$ samples suffice for the cross-style and cross-condition signal we report and pass manual quality assurance on every editing-task sample, but the benchmark remains small relative to training corpora and leans toward academic and poster figures, with infographics the smallest pool; broadening infographic coverage is the natural next iteration.
\section{Case studies}
\label{app:cases}

\begin{figure}[!htbp]
\centering
\includegraphics[width=0.95\linewidth]{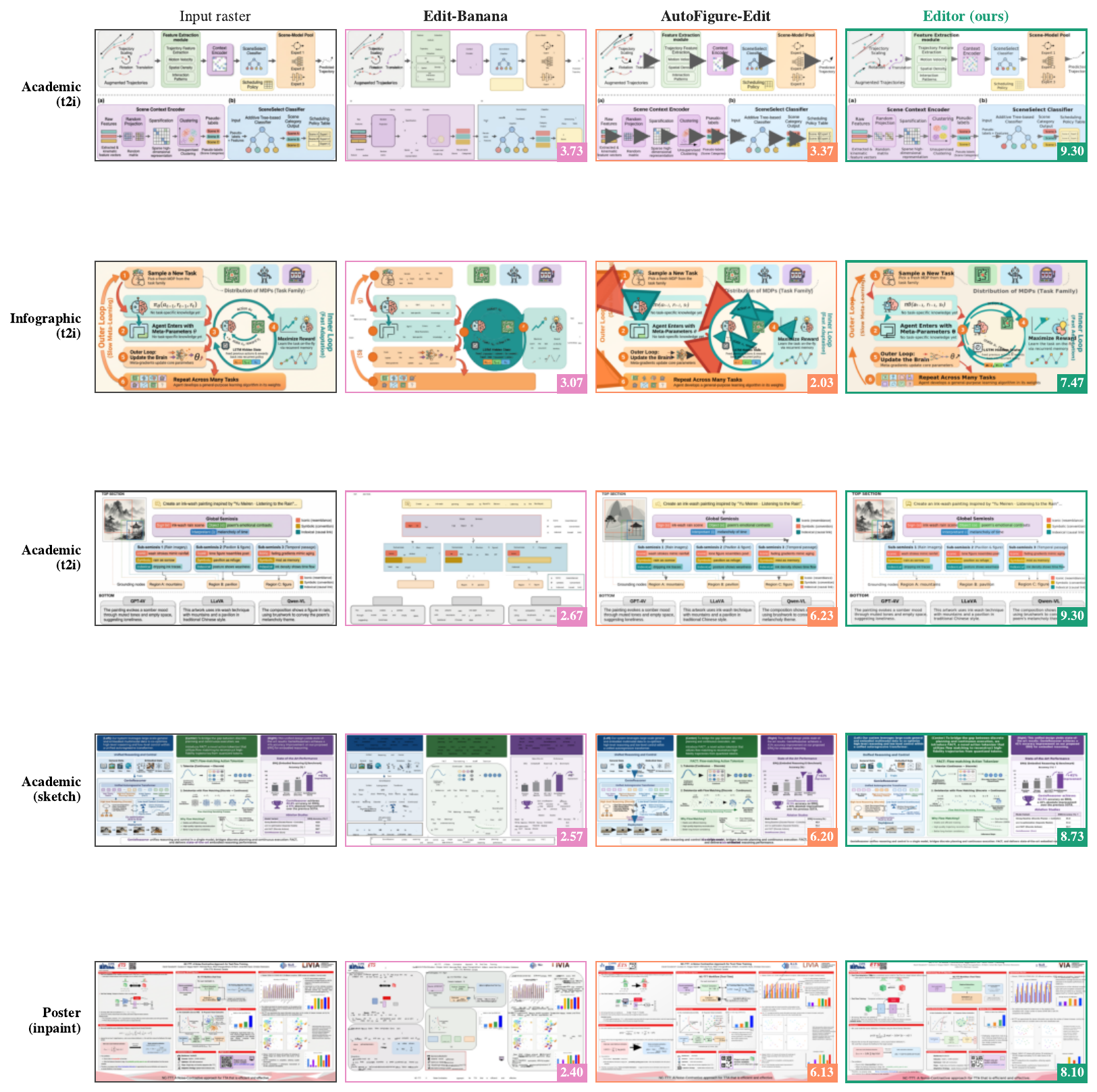}
\caption{Qualitative comparison of editable-output systems on five
cases. Columns: input raster, \EditBanana, \AutoFigureEdit, \Editor\
(green frame). Per-panel numbers are three-VLM judge means.}
\label{fig:editor-qualitative} 
\end{figure}

This appendix collects the qualitative case studies referenced from
the main text. Figure~\ref{fig:editor-qualitative} compares \Editor\
against prior raster-to-editable systems on five \CrafterBench\
outputs; per-panel scores are the three-VLM judge mean.

\noindent\textbf{\Crafter\ on input-honoring editing tasks.}\
On the \CrafterBench\ editing samples shown in
Figure~\ref{fig:conditioning-qualitative}, the baselines visibly
regenerate from the caption and ignore the conditioning input, so the
input-fidelity aspect scores them far below \Crafter. The
conditioning-input columns and the \Crafter\ columns share spatial
structure, while the baseline columns share content with neither.

\noindent\textbf{\Crafter\ on multi-component academic figures.}\
On academic samples that combine multiple paper-named
components and complex flow (e.g., a manipulation-order framework or
a mesh-movement network), the baselines miss the specific
paper-named components and substitute generic stock visuals;
\Crafter's verify-then-refine loop catches the missing component on
the first round, and the structured corrective layer adds a layout
note that pins the component to the correct sub-region for the
second round.

\section{Failure cases}
\label{app:failures}

\begin{figure}[!htbp]
\centering
\includegraphics[width=0.82\linewidth]{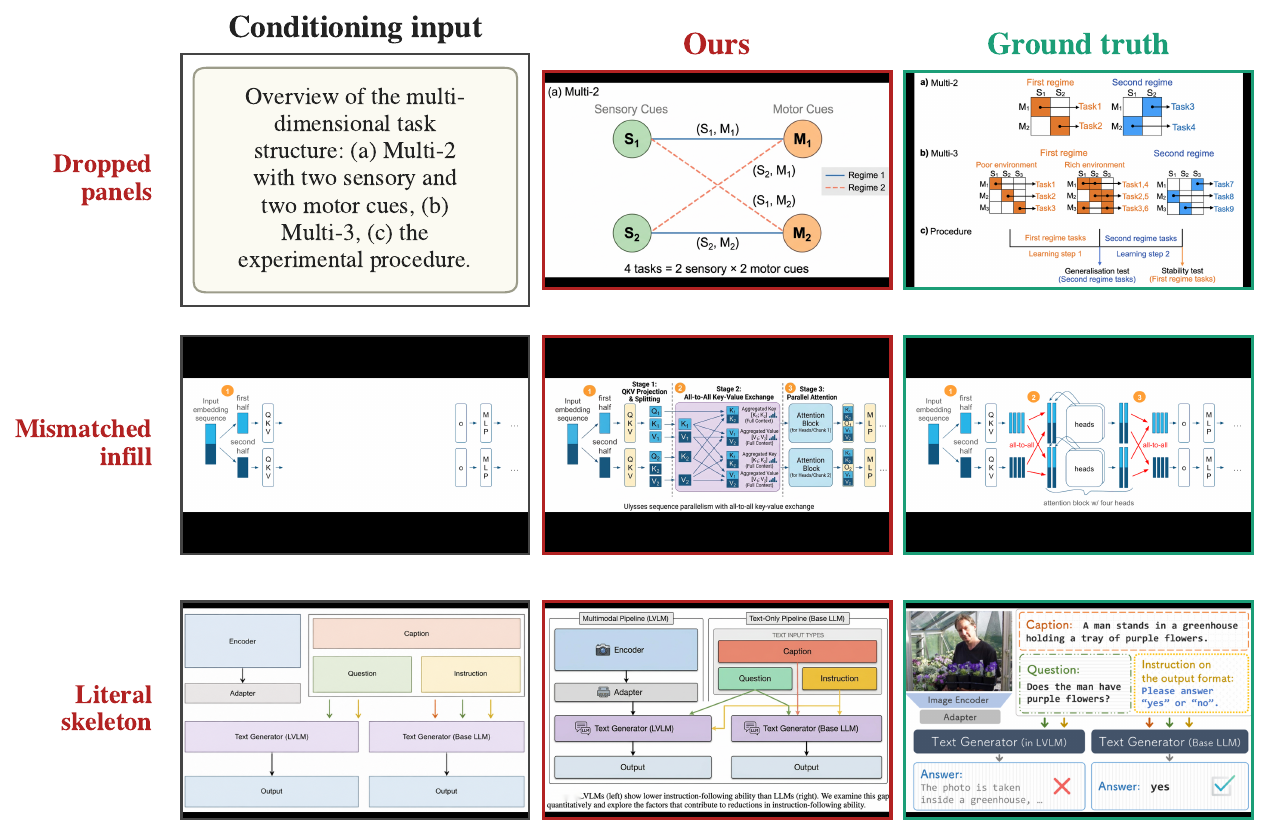}
\caption{Representative \Crafter\ failure cases. Columns:
conditioning input (caption for the text-to-image rows, reference image
for the reference-conditioned rows), \Crafter\ (red frame), ground truth
(green frame). Rows, top to bottom: dropped panels, mismatched infill,
literal skeleton.}
\label{fig:failure-cases}
\end{figure}

\Crafter\ does not win on every sample. Figure~\ref{fig:failure-cases}
collects three cases on which the three-VLM judge prefers the human-drawn
target, in the layout of Figure~\ref{fig:conditioning-qualitative} but
without baseline columns. Each row isolates one failure mode and the
harness stage responsible for it.

\begin{itemize}[leftmargin=1.4em,itemsep=2pt,topsep=2pt]
\item \textbf{Dropped panels} (text-to-image). The intent reasoner
collapses a multi-panel caption (a/b/c) into a single panel; the missing
panels never enter the shared specification, so the verify-then-refine
loop cannot recover them.

\item \textbf{Mismatched infill} (mask completion). \Crafter\ regenerates
the masked region in a clashing boxed register that no longer continues
the surrounding diagram preserved from the input, breaking visual
continuity at the mask boundary.

\item \textbf{Literal skeleton} (sketch-conditioned). The layout follows
the sketch faithfully but stays an abstract block diagram, omitting the
concrete illustrative example (a photo and a worked question/answer) that
the target uses to convey the point.
\end{itemize}

\noindent The first mode traces to intent reasoning (a collapsed panel
count), while the other two trace to the backbone and to a critic that
scores structure but not whether infilled content stays faithful to the
input; both point to concrete fixes: a panel-count check and a
mask-boundary continuity check in the critic.

\newpage

\end{document}